\documentclass[preprint,12pt]{elsarticle}
\makeatletter
\def\ps@pprintTitle{%
  \let\@oddhead\@empty\let\@evenhead\@empty
  \def\@oddfoot{\reset@font\hfil\thepage\hfil}%
  \let\@evenfoot\@oddfoot}
\makeatother

\usepackage{amssymb}
\usepackage{amsmath}

\usepackage{booktabs, subfigure, color}
\usepackage{amsmath, amssymb, amsthm, mathtools}
\usepackage{enumitem, listings}
\usepackage{algorithmic, algorithm}
\usepackage{multicol}

\def\be{\begin{equation}}
\def\ee{\end{equation}}
\def\d{{\text{d}}}

\begin{document}

\begin{frontmatter}



\title{RoboLDA: A Probabilistic Generative Model for Uncovering Embodied Hierarchical Structures in Voxel-based Soft Robots} 


 \author[1]{Junru Song\fnref{fn1}}
 \affiliation[1]{organization={School of Computer Science, Shanghai Jiao Tong University},
             city={Shanghai},
             postcode={200240},
             country={China}}

 \author[2]{Yang Yang\fnref{fn1}}
 \affiliation[2]{organization={Intelligent Game and Decision Laboratory},
             city={Beijing},
             postcode={100071},
             country={China}}

 \author[3]{Jingdan Shi}
 \affiliation[3]{organization={School of Statistics, Renmin University of China},
             city={Beijing},
             postcode={100872},
             country={China}}

 \author[2]{Guozhen Li}
 \author[2]{Weien Zhou}

 \author[4,5]{\newline Ying Wen}
 \affiliation[4]{organization={School of Artificial Intelligence, Shanghai Jiao Tong University},
             city={Shanghai},
             postcode={200030},
             country={China}}
 \affiliation[5]{organization={Shanghai Innovation Institute},
             city={Shanghai},
             country={China}}

 \author[6,3]{Feifei Wang\corref{cor1}}
 \ead{feifei.wang@ruc.edu.cn}
 \affiliation[6]{organization={Center for Applied Statistics, Renmin University of China},
             city={Beijing},
             postcode={100872},
             country={China}}
 \author[2]{Wen Yao\corref{cor1}}  
 \ead{wendy0782@126.com}

 \author[2]{Tingsong Jiang}

 \cortext[cor1]{Corresponding authors.}
 \fntext[fn1]{Co-first authors with equal contribution.}

\begin{abstract}
Recent advances in robotics highlight hierarchical configurations of robot morphology, where multiple levels of functional substructures synergize to facilitate intelligent behaviors. This hierarchical perspective, while particularly advantageous for voxel-based soft robots (VSRs) to ease design and control complexities, is hindered by its heavy reliance on domain expertise. In this work, we address the following question: can we derive such hierarchical design principles solely from existing successful designs? We answer affirmatively by presenting RoboLDA, a Bayesian probabilistic model that decomposes VSR morphology generation into a four-level hierarchy: ``task-robot-organ-voxel", and is trained via variational inference. Through extensive experiments on simulated VSRs, we verify the presence of consistent, intuitive hierarchical patterns underlying high-performing VSR designs and showcase RoboLDA's proficiency to extract and leverage these hierarchical priors for \emph{zero-shot} robot design in unseen tasks. The generated designs, even without further optimization, achieve on average 106.4\% of the optimized performance produced by evolutionary algorithms. Additionally, the organ structures inferred by RoboLDA serve as valid functional substructures, significantly enhancing synergistic motion control when integrated with modular control policies. Our work pioneers hierarchical generative modeling of robot morphology, offering a promising pathway towards more interpretable and generalizable development of embodied agents. Our code and data are open-sourced to support future research\footnote{Code and data available at: https://anonymous.4open.science/r/RoboLDA}.
\end{abstract}




\end{frontmatter}



\section{Introduction}
\label{submission}

Biological systems are characterized by hierarchical organization, where simple building blocks assemble into intermediate functional substructures that collectively enable complex behaviors. 
Such hierarchical decomposition has been widely regarded as an effective mechanism for balancing structural complexity, specialization, and robustness \citep{parker2024organ,baluvska2024sentient,gottlieb2025gene}. Motivated by these observations, 
the concept of functional substructures within hierarchical robotic systems has been highlighted by a growing number of studies. For example, numerous studies on high-dimensional robotic control establish the benefit of muscle synergies for designing low-complexity yet effective control algorithms \cite{chen2019realizing,dong2022low,chen2024robust}. Meanwhile, a variety of self-reconfigurable modular robots have been devised, achieving unprecedented versatility and scalability in design and motion control \cite{christensen2010anatomy,meng2011autonomous,dai2024research,fang2025hierarchically}. While these studies collectively suggest a promising vision of hierarchical morphological design, there still lacks research that mathematically formalizes such hierarchical structures within robot morphology. In this respect, \citet{howard2019evolving} is particularly noteworthy, as they offer a radical perspective on the convergence of materials, manufacturing, and design towards a task-oriented hierarchical scheme for morphological evolution. However, this perspective lacks mathematical formalization and experimental validation, leaving the feasibility and practicality of a hierarchical robot design framework largely unexplored. 

Recently, voxel-based soft robots (VSRs) have garnered extensive attention due to their distinct advantages, including adaptability, biomimetic properties, and suitability for human-robot interaction \cite{hiller2011automatic,rossiter2021soft,legrand2023reconfigurable}. 
A VSR is a modular soft robot represented as a grid of discrete, interconnected volumetric elements (voxels). Different voxels can encode different material and actuation properties, and their combination yields a flexible and expressive body representation that is particularly suitable for morphology design and analysis.
However, these advantages are accompanied by several major challenges, such as their combinatorially vast and highly abstract design spaces, as well as high degrees of freedom in soft materials. While numerous modular or hierarchical soft robotic systems have been devised to ease the complexities in design and control \cite{zhang2020modular,wu2024modular,fang2025hierarchically}, their design principles rely heavily on domain expertise and are largely incompatible with one another. To our knowledge, no universal framework currently exists that characterizes the hierarchical configurations of VSRs in a general sense. There have also been very limited attempts to derive insights from the abundant successful designs that already exist — whether manually designed or automatically evolved — to facilitate more accessible and interpretable design processes \citep{saito2024effective,apraez2025morphological}. 

\begin{figure}[h!]  
    \centering  
    \includegraphics[width=0.8\textwidth]{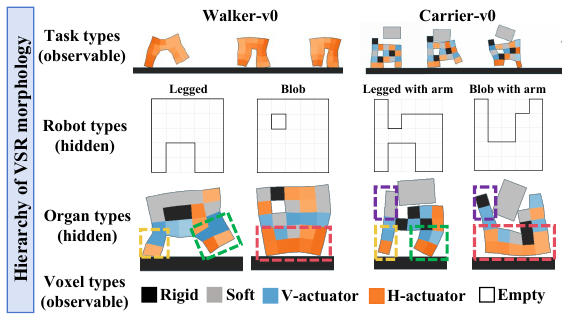}  
    \caption{VSR hierarchy from tasks to voxels. Robot types specialized for different tasks exhibit distinct features, but also share some common functional substructures (colored dashed boxes). ``V" and ``H" represent ``vertical" and ``horizontal", respectively. }
    \label{fig:VSR-examples}  
\end{figure} 

To this end, we introduce RoboLDA, a probabilistic generative model that infers embodied hierarchical structures from voxel-based soft robot designs. 
As illustrated in Figure \ref{fig:VSR-examples}, VSR morphologies can be naturally organized into a four-level hierarchy--\textbf{task, robot, organ, and voxel}--where different task types (top row) give rise to distinct high-level robot types (\emph{e.g.}, \emph{legged} versus \emph{blob}), while sharing recurring substructures. Conditioned on task types, RoboLDA models each morphology as arising from latent robot types, which determine spatial layouts of organ types corresponding to functional substructures (highlighted by colored dashed boxes in Figure \ref{fig:VSR-examples}). These organ layouts are then instantiated by heterogeneous voxel types (bottom row) to form complete robot bodies.
This generative process shares the spirit of Latent Dirichlet Allocation (LDA) \citep{blei2003latent} for discrete data modeling, enabling the discovery of hidden correlations within ``robot-organ'' and ``organ-voxel'' substructures. Meanwhile, we substantially relax the ``bag-of-words" assumption of LDA using meticulously designed model architectures to account for the multilayered, spatial relationships essential in robot morphology. We then present an efficient approach for training RoboLDA using the Variational Autoencoder \cite{kingma2013auto}, and also show how RoboLDA could transfer hierarchical design principles to unseen tasks in a \emph{zero-shot} manner. Additionally, we showcase that the organ structures identified by RoboLDA through variational inference can also benefit broader downstream applications, such as being integrated into modular control algorithms for synergistic motion control. 

To justify the architectural design of RoboLDA and its effectiveness to uncover hierarchical structures from VSR morphologies, we conduct a suite of carefully designed simulated experiments. We start by pre-training RoboLDA on pre-collected high-performing VSRs from diverse task settings and reveal the presence of consistent and interpretable hierarchical patterns. We then transfer the learned design principles to new tasks for morphology generation. RoboLDA exhibits impressive \emph{zero-shot} generalizability, achieving an average of 106.4\% of the optimal performance obtained by evolutionary algorithms. Additionally, we show that the organs inferred by RoboLDA can also serve superiorly as functional substructures. By enhancing intra-organ message passing within a Transformer-based modular control policy, RoboLDA promotes more synergistic and dexterous motion control, outperforming state-of-the-art baselines. 

Importantly, unlike prior work that focuses on iterative morphology--control co-optimization \citep{hu2022modular,liu2025morphology,song2024morphvae,song2025laser}, we study an orthogonal problem: how to model, interpret, and reuse hierarchical structural priors from existing morphologies. RoboLDA is accordingly formulated as an offline prior learner, though it can naturally serve as an initialization for closed-loop optimization via estimation-of-distribution algorithms (EDAs). Our evaluation is conducted in simulation; real-hardware transfer is discussed as future work.

Our contributions are as follows: (a) We reveal the existence of hierarchical generative processes governing high-performing soft robot morphologies. (b) We introduce RoboLDA, a pioneering probabilistic model to deduce multilayered design principles from morphology samples, with minimal reliance on domain expertise. (c) Through experiments we showcase RoboLDA's versatility to benefit both robot design and control. Its generic formulation is also shown to apply beyond VSRs (\emph{e.g.}, articulated rigid robots), opening up possibilities of broader applications and future research. 

\section{Related Work}
\label{sec:related work}

\subsection{Hierarchical robotic structures}

Multi-cellular organisms in nature generally exhibit hierarchical structures, where cells first assemble to be various levels of functional substructures like organs and muscle synergies before forming an entire body \cite{parker2024organ,baluvska2024sentient,gottlieb2025gene}. Learning the lesson from nature, robotic researchers leverage such concepts to devise modular robots \cite{christensen2010anatomy,meng2011autonomous,dai2024research,fang2025hierarchically}, synergistic control algorithms \cite{chen2019realizing,dong2022low,chen2024robust,stella2025synergy}, etc., yielding promising results. Recent studies analyze task-dependent structural specialization in VSRs, revealing that distinct tasks induce characteristic morphological patterns \citep{ferigo2025totipotent}. Along a related line, \citet{nadizar2025enhancing} demonstrates that maintaining diversity across multiple levels can substantially enhance adaptability and zero-shot performance in soft robots. While the significance of functional substructures in robots is substantiated, it remains unexplored whether such hierarchical design experience could be automatically induced from various high-performing robot morphologies and subsequently leveraged for designing new morphologies, even in unseen tasks. In this respect, the most relevant study prior to us is \citet{howard2019evolving}, which introduces a hierarchical architecture for morphological representation. Specifically, \citet{howard2019evolving} present a three-level hierarchy spanning from materials, to components, and to robots. Nevertheless, this approach lacks empirical validation and is only poised for the evolutionary search of robot morphology, instead of mining for hierarchical structures from existing morphologies. On the contrary, we are the first to adopt a probabilistic generative approach towards hierarchical morphology representation, which naturally lends itself to this purpose. 

\subsection{Representation learning of robot morphology}
Our work is partly related to robot design automation, a subdomain in Robotics that lays its emphasis on designing high-performing robot morphology without human intervention. Evolutionary algorithms (EAs) featuring elitism selection and stochastic search operators have been successfully applied to this problem. Among many others, \citet{leger2012darwin2k} pioneers the application of EAs to the design of rigid robots. \citet{wang2019neural} proposes an evolutionary strategy specifically for the search of graph-shaped robot structures. However, these studies all represent robots as their original forms like kinematic trees or material matrices. Such direct encoding has been proven to scale poorly to the vast combinatorial search spaces of modular robots, such as VSRs. Subsequent studies including \citet{auerbach2010dynamic}, \citet{cheney2014unshackling} and \citet{mertan2025controller} demonstrate the potential of Compositional Pattern Producing Networks (CPPNs), as a form of generative encoding, to evolve complex morphologies. Recently, \citet{hu2023glso}, \citet{song2024morphvae}, \citet{li2024generating} and \citet{liu2025morphology} further utilize deep generative models to learn latent representations of robot morphology. \citet{zhao2025cross} explore \emph{zero-shot} generalization of morphology and control in VSRs, by directly transferring optimized designs to new tasks \citep{nadizar2025enhancing}. 
Despite their effectiveness, the aforementioned approaches do not explicitly model hierarchical assembly as an inductive bias. In contrast, RoboLDA introduces a hierarchical generative formulation that characterizes and enables the transfer of anatomical structures across morphologies from different tasks.

We note that our primary objective is \emph{zero-shot} and \emph{interpretable} morphology modeling, which is complementary to the optimization-oriented studies above. RoboLDA can be integrated with EDAs \citep{bhattacharjee2019estimation} or differentiable simulation \citep{hu2019difftaichi} to further refine the generated designs, which we leave for future work.

\section{Method}
\label{sec:method}
\subsection{Preliminary: Latent Dirichlet Allocation}

Our model draws inspiration from Latent Dirichlet Allocation (LDA) \citep{blei2003latent}, a classical probabilistic topic model originally developed for text analysis. In LDA, each document is modeled as a mixture of $K$ latent topics, and each topic $k$ defines a distribution $\varphi_k$ over a vocabulary of words. The generative process for a document is as follows: (i) draw a document-level topic proportion $\vartheta \sim \text{Dir}(\alpha)$; (ii) for each word position $n$, sample a topic assignment $z_n \sim \text{Multi}(\vartheta)$; and (iii) sample the observed word $w_n \sim \text{Multi}(\varphi_{z_n})$. The key analogy motivating RoboLDA is: \emph{robot types} correspond to \emph{documents}, \emph{organs} to \emph{topics}, and \emph{voxels} to \emph{words}. Just as LDA discovers latent thematic structures from a corpus without supervision, RoboLDA discovers latent anatomical structures from a collection of morphologies. However, classical LDA neglects spatial structure, lacks a task-conditioned hierarchy, and treats each document independently. These limitations motivate our extension into a hierarchical probabilistic model tailored to robot morphology.

\subsection{RoboLDA}

\begin{figure*}[ht]
    \centering
    \includegraphics[width=\linewidth]{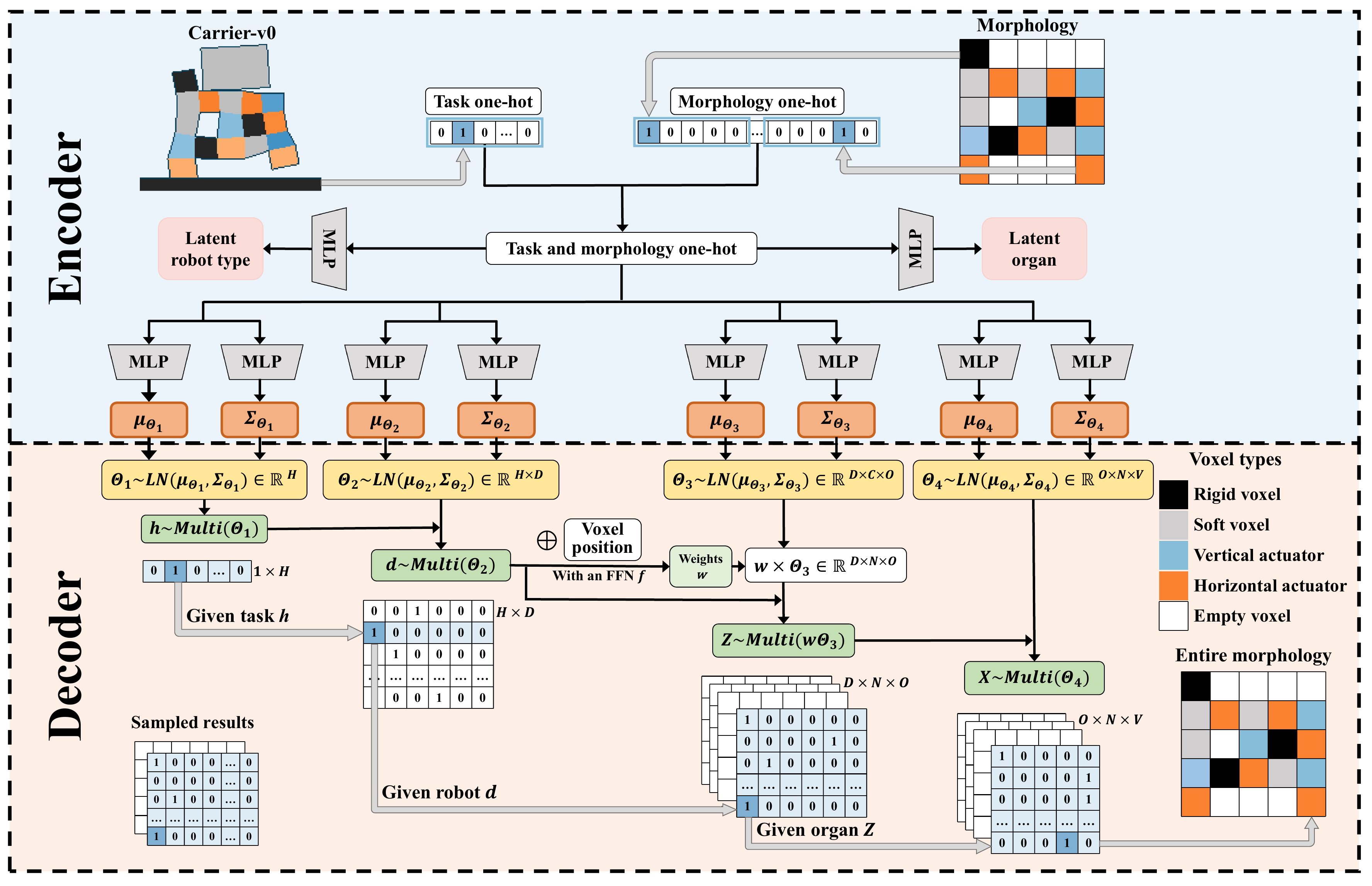}
    \caption{The architecture of RoboLDA, including a decoder (generative process) and an encoder (approximate posterior). The generative process consists of four levels namely task, robot, organ and voxel. The approximate posterior infers latent variables to support variational inference. }
    \label{fig:RoboLDA generation}
\end{figure*}

RoboLDA is a four-level hierarchical probabilistic generative model, characterizing the generative process of VSR morphology as ``task-robot-organ-voxel”. With Variational Autoencoder (VAE) \citep{kingma2013auto} as the basic architecture, our introduction is divided into two parts: generative process and approximate inference. As illustrated in the lower right corner of Figure \ref{fig:RoboLDA generation}, VSRs consist of multiple types of building blocks (known as \emph{voxels}) interconnected in a grid-like layout, and achieve motion control through volumetric actuation. We adopt VSRs simulated by Evolution Gym \cite{bhatia2021evolution} due to its fast simulation support and extensive task suite. However, our mathematical formulation of robot morphology is generalizable to other types of robots; Please see Section \ref{sec:exp} for more details. 

We first introduce some notations. Assume there are $H$ types of tasks, $D$ types of robots, $O$ types of organs and $V$ types of voxels. Each VSR morphology can be represented by a 2D voxel matrix, denoted as $X$. Each entry of $X$ stands for the voxel type at the corresponding position. Denote the height and width of a VSR as $a$ and $b$ respectively, and the total number of voxels in a VSR as $N=a\times b$. Denote Dirichlet and multinomial distributions as ``Dir" and ``Multi", respectively. Let $\Theta_{i,j,k}$ denote the $j$-th slice along the first dimension and $k$-th slice along the second dimension of multi-dimensional array $\Theta_i$. 

We extend LDA's two-level generative formulation (document--topic--word) to a four-level hierarchy (task--robot--organ--voxel) and substantially relax its ``bag-of-words" assumption to accommodate spatial information in robot morphology. The generative process of RoboLDA, shown in the lower half of Figure \ref{fig:RoboLDA generation}, is decomposed into the following steps. In Figure \ref{fig:RoboLDA generation}, the orange boxes correspond to global latent distributions, while the green boxes indicate per-robot sampling steps.

\begin{itemize}
    \item[1.] Generate task type probabilities $\Theta_1\sim \text{Dir}(\alpha)$ over all $H$ tasks. 
    \item[2.] For each task $h$, generate robot type probabilities $\Theta_{2,h}\sim\text{Dir}(\beta)$. (These correspond to the first two orange latent boxes in the decoder, and their sampled counterparts are the green ``Given task $h$'' and ``Given robot $d$'' boxes.)
    \item[3.] For each robot type $d$, generate organ type probabilities for all voxel positions. (Analogous to the topic proportion $\vartheta$ in LDA; corresponds to the middle decoder module in Figure \ref{fig:RoboLDA generation}.) To avoid parameter proliferation, here organ distributions across different positions are defined as weighted combinations of $C$ \emph{``organ components"} with $C<N$. Generate each of these components: $\Theta_{3,d,c}\sim\text{Dir}(\gamma)$ for $c\in\{1,\cdots,C\}$. 
    \item[4.] For each organ $z$, generate voxel type probabilities for all positions: $\Theta_{4,z,n}\sim\text{Dir}(\mu)$ for $n\in\{1,\cdots,N\}$. (Analogous to the topic--word distribution $\varphi_k$ in LDA; corresponds to the rightmost orange latent box in Figure \ref{fig:RoboLDA generation}.)
    \item[] Steps 1--4 define the orange global latents in Figure \ref{fig:RoboLDA generation}; Step 5 is the green per-robot sampling path, where the 0--1 matrices are one-hot encodings at each level.
    \item[5.] For each robot $X$:
    \begin{itemize}[leftmargin=0.8em]
        \item[(a)] Choose a task type $h$ with probabilities given by $\Theta_1$: $h\sim \text{Multi}(\Theta_1)$. 
        \item[(b)] Choose a robot type $d$ with probabilities $\Theta_{2,h}$: $d\sim \text{Multi}(\Theta_{2,h})$. 
        \item[(c)] For the $n$-th voxel in robot $X$, $n\in\{1,\cdots,N\}$: 
        \begin{itemize}[leftmargin=0.8em]
            \item[i.] Calculate weights of organ components: $w_n \leftarrow f(h,p_n)$, where $f$ is an MLP (three 128-unit hidden layers, $\tanh$ activations) mapping the concatenation of task type $h$ and voxel coordinates $p_n$ to the organ-component weight vector. Choose an organ type $Z_n$ with probabilities given by $w_n\cdot \Theta_{3,d}$: $Z_n\sim\text{Multi}(w_n\cdot \Theta_{3,d})$. (Analogous to the per-word topic assignment $z_n\sim\text{Multi}(\vartheta)$ in LDA, but here the organ distribution is spatially varying rather than uniform across positions.) Denote the organs at all positions collectively as $Z$, which is referred to as the \emph{``organ layout"}. 
            \item[ii.] Choose a voxel type $X_n$ with probabilities given by $\Theta_{4,Z_n,n}$: $X_n\sim\text{Multi}(\Theta_{4,Z_n,n})$. (Analogous to the per-word generation $w_n\sim\text{Multi}(\varphi_{z_n})$ in LDA.)
        \end{itemize}
    \end{itemize}
\end{itemize} 

In the above generative process, all distribution parameters $\Theta_i$ ($i=1,\cdots,4$) are ``global" latent variables as they govern all robots. Following \citet{srivastava2017autoencoding}, we approximate their Dirichlet prior distributions with a Logistic Normal (LN) distribution due to its convenience for reparametrization \cite{williams1992simple}. The robot type $d$ and organs $Z$ of each robot are ``local" latent variables as they are independent between different robots. 

Notably, RoboLDA's core contribution is \emph{vertical} hierarchical decomposition along the ``task--robot--organ--voxel'' axis, which is largely orthogonal to the \emph{horizontal} message passing among voxels addressed by GNN and Transformer-based approaches \citep{wang2018nervenet,gupta2022metamorph,hao2024heteromorpheus}. As a general probabilistic framework, RoboLDA is in principle compatible with different neural architectures. We opt for MLP-based components in the current implementation to isolate the contribution of hierarchical modeling without confounding architectural factors. Integrating GNN-based spatial reasoning into RoboLDA is a promising direction for future work.

\subsection{Variational inference of RoboLDA}

The training of RoboLDA and inference of latent variables both resort to variational inference. The architecture of RoboLDA, including encoder (approximate posterior) and decoder (generative process), is illustrated in Figure \ref{fig:RoboLDA generation}. In particular, the upper half of Figure \ref{fig:RoboLDA generation} corresponds to the encoder, where task and morphology one-hot inputs are processed by MLPs to produce the approximate posterior parameters of the global latent variables. Denote high-performing morphologies as $\tilde{X}$, and their corresponding tasks as $\tilde{T}$. To conduct variational inference, we construct approximate posteriors for global latents as: 
\begin{equation}
\label{eq:pos}
    \mu_{\Theta_i} = \phi_{\mu_{\Theta_i}}(\tilde{X},\tilde{H}), \quad
    \Sigma_{\Theta_i} = \phi_{\Sigma_{\Theta_i}}(\tilde{X},\tilde{H}), \quad
    \Theta_i\sim LN(\mu_{\Theta_i},\Sigma_{\Theta_i})
\end{equation}

\noindent for $i=1,2,3,4$, where $\mu_{\Theta_i}$ and $\Sigma_{\Theta_i}$ are fully-connected networks, and $\tilde{X}$ and $\tilde{H}$ are both one-hot encoded and flattened to one dimension beforehand. 

The robot type and organs of each morphology is then inferred with the following approximate posteriors: 
\begin{equation}
\label{eq:species}
p_d = \phi_d(X,h,\Theta_2), d\sim \text{Multi}(\text{Softmax}(p_d)),
\end{equation}
\begin{equation}
\label{eq:organ}
p_z = \phi_z(X,h,\Theta_3), Z\sim \text{Multi}(\text{Softmax}(p_z)),
\end{equation}
where $\phi_d$ and $\phi_z$ are fully-connected networks that take the robot morphology $X$, task type $h$, along with inferred global latents (omitted in Figure \ref{fig:RoboLDA generation}), as input. 

Based on the above approximate posteriors, an evidence lower bound (ELBO) is constructed as the objective function, and the trainable parameters in RoboLDA, \emph{i.e.,} those in fully-connected layers, are optimized via gradient ascent. The approximate posterior in Eq.(\ref{eq:organ}) is also used to infer organ structures from robot morphologies, which can be leveraged for various downstream applications such as robotic control. Please see \ref{ELBO} for the detailed derivation of ELBO, and find more systematic introduction to VAE in \citet{kingma2013auto}. 

\subsection{Downstream applications of RoboLDA}
\subsubsection{Zero-shot robot design with RoboLDA}
\label{sec:zero-shot}

Since the generative process of high-performing robot morphologies is explicitly modeled, we can directly transfer the design experience from existing tasks to new ones, by leveraging the inferred global latents. Specifically, to generate robot morphology for an unseen task, we start with the second layer of the hierarchy and uniformly sample a robot type. The approximate posteriors of $\Theta_3$ and $\Theta_4$ are then taken as their new priors to produce organs and voxels. In Section \ref{sec:ea} we confirm that RoboLDA can indeed effectively use prior knowledge extracted in this manner to achieve zero-shot robot proposal. Here ``zero-shot" indicates that RoboLDA relies solely on robot designs from training tasks without further optimizing the generated robots in new tasks. 

\subsubsection{Robot control with organ synergy}

This section details how the organ structures derived from RoboLDA can be used for synergistic control. We employ MetaMorph \cite{gupta2022metamorph}, a Transformer-based modular control policy, as our base controller. Recall that the organ layout $Z$ of each robot can be inferred according to Eq.(\ref{eq:organ}). We leverage $Z$ as attention masks to regulate message passing and enhance \emph{intra-organ} (\emph{i.e.,} among voxels that belong to the same organ) communication in the controller. By integrating MetaMorph with organ structures, we are able to learn a universal controller for different morphologies while allowing each of them to have heterogeneous synergy patterns among voxels.

\begin{figure*}[ht]  
    \centering  
    \includegraphics[width=\textwidth]{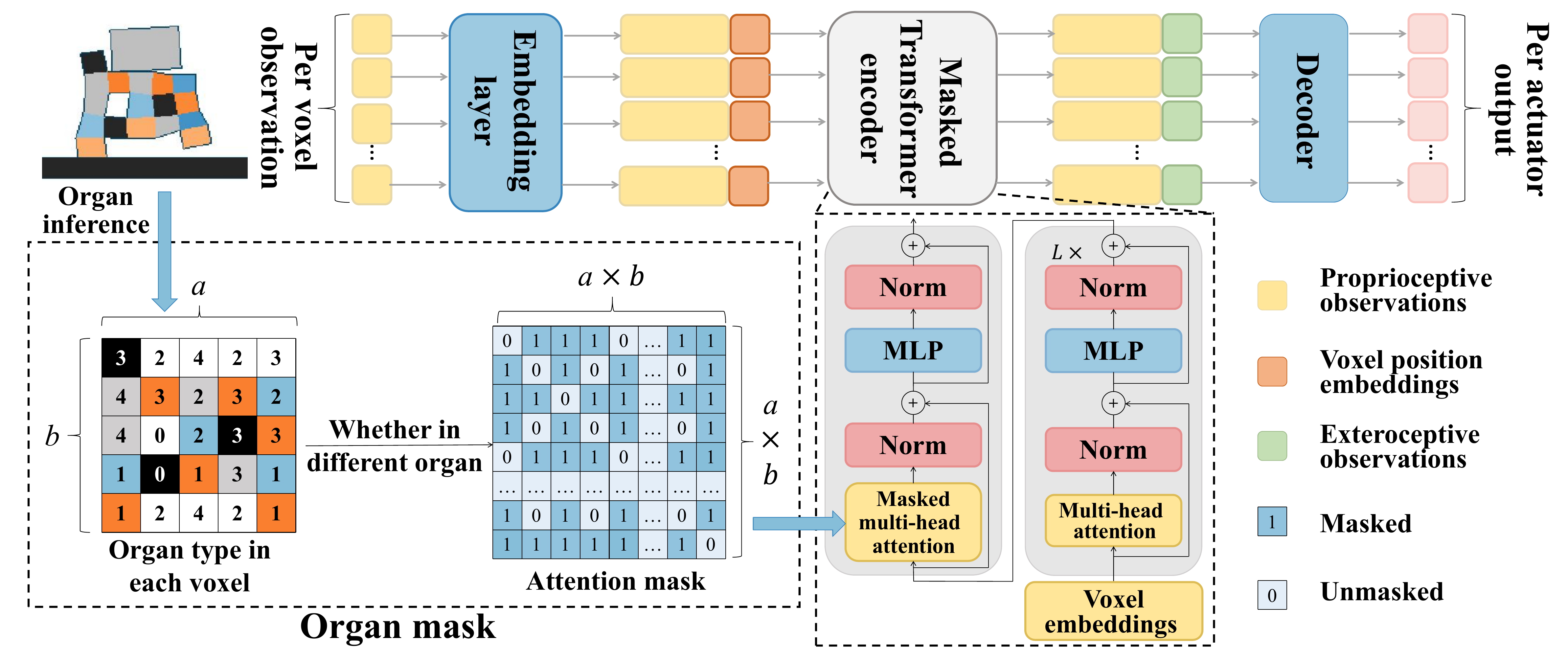} 
    \caption{MetaMorph with organ synergy. The upper half shows the architecture of MetaMorph. The lower half demonstrates how organs are converted into attention masks and imposed onto the attention layer. }  
    \label{fig:MetaMorph+organ}  
\end{figure*}

Figure \ref{fig:MetaMorph+organ} illustrates the architecture of MetaMorph. MetaMorph treats each voxel as a separate token, and takes the proprioceptive observations, $s_l$, of all the voxels as input. The $s_l$'s are then linearly embedded and added with learnable positional embeddings $W_{pos}$ before being fed into a Transformer encoder consisting of five self-attention layers:
\begin{equation}
\label{eq:pi1}
m_0 = \phi_1(s_l)+W_{pos},
\end{equation}
\begin{equation}
\label{eq:pi2}
m_k = SA(m_{k-1}), k=1, \cdots, 5, 
\end{equation}
where $\phi_1$ denotes linear projection and $SA$ is a multi-head self-attention layer \cite{vaswani2017attention}. The output of the Transformer is concatenated with linearly projected exteroceptive observations ($s_g$), before being decoded into action signals: 
\begin{equation}
\label{eq:pi3}
a = \phi_3(m_5, \phi_2(s_g)), 
\end{equation}

\noindent where both $\phi_2$ and $\phi_3$ are linear projection layers. The signals of actuator voxels (\emph{i.e.,} horizontal and vertical actuators) are extracted to control the robot and step the environment into the next state. Note that the above calculation is repeated in each timestep, with subscript $t$ omitted for brevity. Denote Eq.(\ref{eq:pi1}-\ref{eq:pi3}) collectively as $\pi_\psi$ where $\psi$ stands for the trainable parameters, and $(s_l, s_g)$ as $s$. For more details of MetaMorph, the reader is referred to \citet{gupta2022metamorph}. 

We apply organ layout $Z$ inferred by RoboLDA to MetaMorph by first converting it to an attention mask $M$ (Eq.(\ref{eq:mask})). $M$ is then imposed onto the last self-attention layer (\emph{i.e.,} $k=5$ in Eq.(\ref{eq:pi2})) to reinforce intra-organ coordination. 
\vspace{-2.5mm}
\begin{equation}
\label{eq:mask}
M=(M_{i,j}),
M_{i,j}=
\begin{cases}
\text{False},  & \text{if $Z_i = Z_j$, } \\
\text{True}, & \text{if $Z_i \neq Z_j$, }
\end{cases}
\vspace{-1.5mm}
\end{equation}

\noindent for $i,j=1,\cdots,N$, and $Z_i$ represents the organ index of the $i$-th voxel in the robot. $M_{i,j}=\text{True}$ indicates that the voxels $i$ and $j$ do not belong to the same organ, and thus the attention is blocked between them. The training process of MetaMorph assisted with organ synergy is summarized in Algorithm 1 in \ref{metamorph-alg}. 

\subsection{Extension to rigid robots}
\label{sec:extension}

\begin{figure}[h!]
    \centering
    \includegraphics[width=0.5\linewidth]{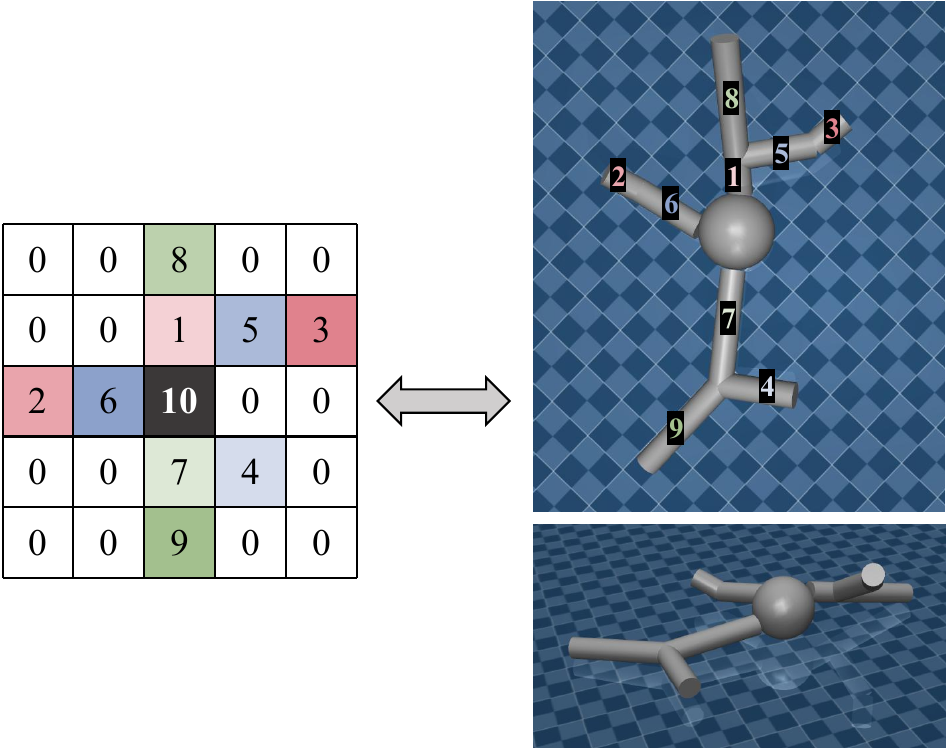}
   \caption{An example of transformation between voxel matrices and rigid robots. For the purpose of illustration, robot joints are torqued so that the directions of rotation become more apparent. Different limb lengths are marked as different colors, while different rotation directions as different shades. Apart from ten different limb types (0$\sim$9), we use a distinct number (10) to denote the torso. }
    \label{fig:example}
\end{figure}

While RoboLDA is initially devised and presented for voxel-based soft robots, here we provide an example of how it could also be extended to other types of robots. We opt for jointed rigid robots, another robot category that is commonly seen in both academia and industry. The extension primarily involves how to construct a reasonable one-on-one mapping between voxel matrices and the kinematic trees of rigid robots. We stick with the $5\times5$ voxel matrix, and assume that the center voxel (\emph{i.e.,} the voxel in the second column and second row) is always a torso. The remaining voxels correspond to limbs that are connected either to the torso or to one another via joints according to the adjacency relations. In addition to the torso, we introduce ten voxel types: one for empty voxel, and nine for limbs with varying lengths and directions of rotation. Please see Figure \ref{fig:example} for a concrete example of transformation between voxel matrices and rigid robots. For illustration we only assign one degree of freedom to each joint (\emph{i.e.,} rotate around either $x$, $y$ or $z$ axis), and assume that during resting state, all the limbs are parallel to the $x$-$y$ plane. The design space can be expanded by defining a wider range of voxel types. We provide the complete scripts for procedurally transforming voxel matrices into URDF files, and then into XML files readable by the Mujoco simulation environment \citep{todorov2012mujoco}. The effectiveness of RoboLDA in designing rigid robots is validated in \ref{sec:rigid-validation}. 

\section{Experiments}
\label{sec:exp}
\subsection{Experimental setup}

We base our experiments on Evolution Gym \cite{bhatia2021evolution}, one of the most widely used simulation environments for VSRs. We select nine tasks with varying difficulty levels for benchmarking, including seven locomotion tasks (Walker-v0, BridgeWalker-v0, UpStepper-v0, DownStepper-v0, Climber-v1\&2, GapJumper-v0) and two manipulation tasks (Pusher-v0, Carrier-v0). See Supplementary Material A for more details of the tasks. Following the standard setup in previous research \cite{bhatia2021evolution,wang2023preco,song2024morphvae,liu2025morphology}, both the height and width of VSRs are set as 5; With five types of voxels (as shown in Figure \ref{fig:RoboLDA generation}), this configuration results in around $10^{17}$ possible robot designs, already representing a challenge to design algorithms \citep{mertan2025evolutionary}. For morphology validity, we strictly follow Evolution Gym's default constraints: all voxels must form a single connected component, and each robot must contain at least one actuator voxel (horizontal or vertical). Invalid morphologies are directly skipped, and candidate designs are resampled until the required sample size is reached. 

To evaluate the fitness of robot designs, we employ Proximal Policy Optimization (PPO) \citep{schulman2017proximal} to optimize sensorimotor controllers. PPO is chosen for its training stability and widespread adoption in prior VSR studies \citep{bhatia2021evolution,wang2023curriculum,wang2023preco}. PPO serves here as a standardized evaluation tool rather than a component of RoboLDA. Our preliminary comparison with off-policy alternatives (SAC, TD3) confirmed that PPO provides more stable convergence on this benchmark, likely because the sparse, small-scale rewards make off-policy value estimation less reliable.
Fitness is measured as the cumulative sum of reward achieved over a complete episode in the task environment, using optimized controllers. Please refer to Supplementary Material A for the definition of reward functions and their relationship with actual task performance. In robot design experiments, multilayer perceptrons (MLPs) are utilized as policy networks for simplicity. All the experimental results are averaged across three independent runs. Detailed parameter settings are relegated to Supplementary Material B. Sensitivity analyses regarding random initialization and parameter settings are included in \ref{sec:sensitivity}. Our code and database are available at https://anonymous.4open.science/r/RoboLDA.

\subsection{Collecting high-performing morphologies}
\label{sec:collect}

RoboLDA is trained on high-performing robot morphologies. We note that access to such morphologies is not a strong assumption but a common scenario encountered in practical robot design applications. Specifically, these morphologies could be obtained from both open-source repositories (\emph{e.g.,} those in \citet{gupta2022metamorph, chen2025large,le2025robodesign1m}) and prior design processes. Here we artificially create such a problem setup by evolving morphologies with three prominent evolutionary algorithms -- GA, BO and CPPN-NEAT. The adoption of distinct algorithms, as well as random repeated trials of each, largely ensures comprehensive coverage of the design space. The top 5\% of evaluated designs are selected as high-performing sample. The above EAs are introduced in Supplementary Material C in detail. Note that the robot designs collected in this manner \emph{do not} have any labeled hierarchical structures, leading to a more challenging setup where robot types and organs are to be inferred from data. However, any available domain knowledge regarding pre-defined subpopulations and functional substructures can be conveniently incorporated in the forms of robot types and organs by specifying their prior distributions. 

\subsection{Hierarchical structure inference}

\begin{figure}[ht]  
    \centering  
    \includegraphics[width=\textwidth]{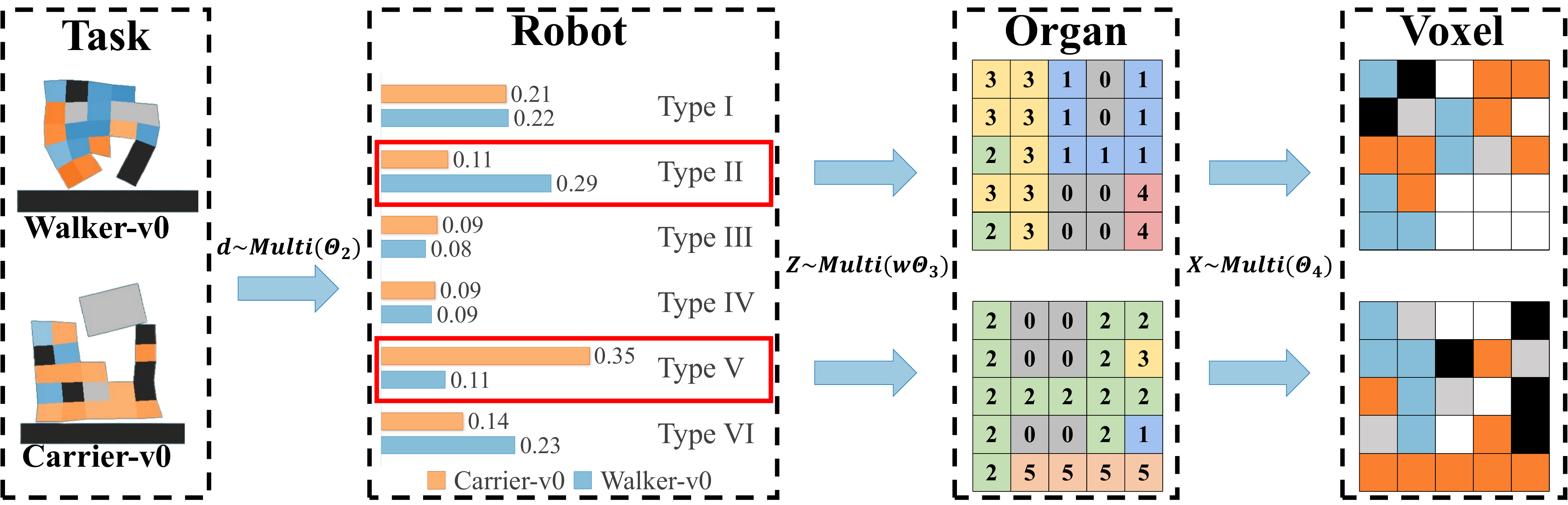} 
    \caption{Illustration of the hierarchical generative process learned by RoboLDA. The distributions of robot type are illustrated as histograms, whereas for organ and voxel layers, each position is labeled with the most likely outcome. }
    \label{fig:structure-result}  
\end{figure}

Here we take two tasks—Walker-v0 and Carrier-v0—as examples and showcase RoboLDA's proficiency in modeling hierarchical structures in multi-task settings. We visualize the learned hierarchical generative process in Figure \ref{fig:structure-result}. It can be observed that in Walker-v0, where robot type II appears with a higher probability, functional substructures resembling legs (represented as organ types 1 and 3) emerge in the organ layer. In Carrier-v0, where robot type V dominates, functional substructures for peristaltic locomotion (organ type 5 in the bottom row) and arm-like structures for lifting (organ type 2) are evident in the organ layer. In the voxel layer, the two robot morphologies derived from the respective organ distributions are displayed: the robot at the top right represents the high-performing morphology for Walker-v0, which is a legged robot, while the robot at the bottom right represents the high-performing morphology for Carrier-v0, which is a peristaltic robot with arms. This is consistent with human-in-the-loop evidence that intrinsic body factors strongly determine task success in modular robots \citep{pietrosanti2023humancontrol}, further suggesting that high-performing morphologies share structural patterns aligned with natural priors. Please find more visualizations of hierarchical structures in Supplementary Material E. Animated GIFs showing these robots interacting with their task environments are available in our anonymous repository, providing an intuitive demonstration of how the inferred functional substructures coordinate during execution.

We note that the hierarchical structures learned by RoboLDA are not always perfectly interpretable, which we believe is a common limitation of data-driven approaches. Nevertheless, given the highly abstract and counter-intuitive nature of voxel-based design spaces \citep{mertan2024investigating}, overemphasizing interpretability would inevitably limit the performance of design algorithms \citep{assis2025performance}. In this sense, we believe RoboLDA strikes a rather satisfactory balance, introducing higher transparency than previous closed-box approaches while achieving competitive performance. It would be a valuable future direction to investigate how human intuition and data-driven insights could be better reconciled. 

\begin{figure}[ht]  
    \centering  
    \includegraphics[width=0.45\textwidth]{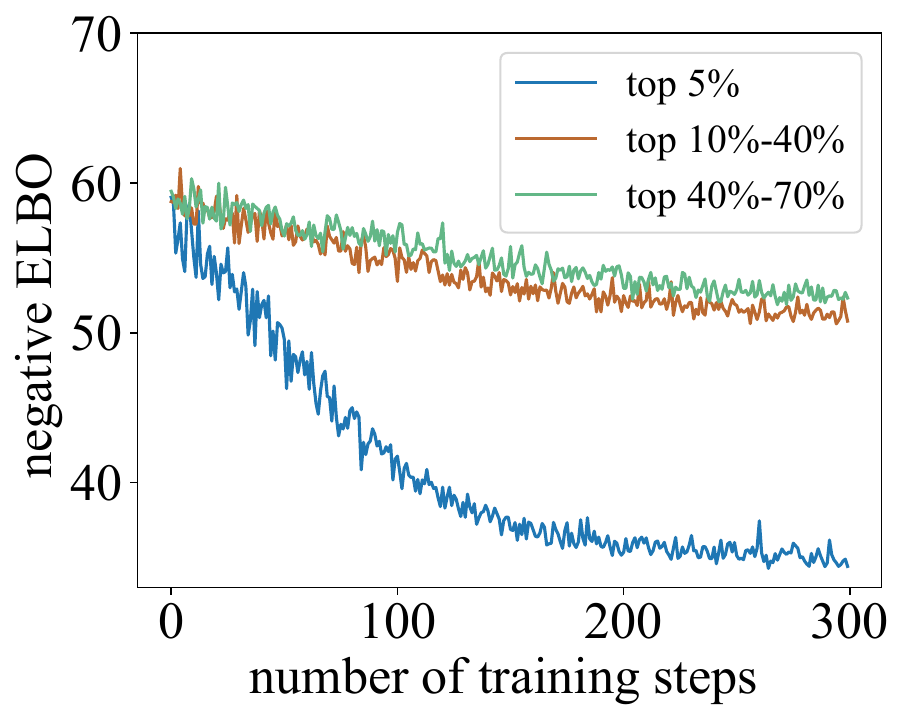} 
    \caption{Learning curves of RoboLDA when being trained on morphologies with different performance levels. }
    \label{fig:sub-optimal}  
\end{figure}

To investigate the learning behaviors of RoboLDA, we additionally substitute the training sample with sub-optimal morphologies. Specifically, we experiment with two alternative performance levels: top 10\%-40\% and top 40\%-70\% among all evolved morphologies described in Section \ref{sec:collect}, and plot the learning curves in Figure \ref{fig:sub-optimal}. It can be seen that training sample with higher performance also leads to notably better goodness-of-fit (lower loss curves). These results highlight the existence of consistent hierarchical patterns within high-performing morphological structures, which is a wealth of knowledge largely underutilized in previous studies. This also justifies RoboLDA's architectural design that echoes the empirical observation. 

\subsection{Zero-shot robot design}
\label{sec:ea}

We proceed to verify whether RoboLDA could generalize the learned hierarchical structures to produce new morphologies. 
For this purpose, RoboLDA is first fitted on a set of pre-training tasks, after which it is queried to generate robot designs for an unseen task. Our choice of train--test task relations is partly inspired by PreCo \citep{wang2023preco}, which considers transfer between structurally related co-design tasks. At the same time, we acknowledge that such transfer settings still involve a degree of human prior about task structure, motivating the quantitative similarity analysis below.
To quantify the relatedness between pre-training and unseen tasks, we measure task similarity by the average edit distance between their collected high-performing morphologies; lower edit distance indicates higher morphological similarity. Based on this metric, the nine transfer settings are ranked and partitioned into three equal-sized groups: high, medium, and low similarity. Table~\ref{tab:task combinations} summarizes the zero-shot transfer settings, together with task annotations and similarity scores. 

\setlength{\tabcolsep}{1mm}
\begin{table}[htb]
\fontsize{8pt}{7pt}\selectfont
\centering
\caption{Task combinations for zero-shot evaluation. 
Each task is annotated with its difficulty level (easy/medium/hard) and task category (loc: locomotion, man: manipulation, cli: climbing). Following the task-relation perspective of PreCo \citep{wang2023preco}, these settings broadly cover three transfer types: more challenging variants (e.g., Climber-v0 $\rightarrow$ Climber-v2), transfers of comparable difficulty (e.g., Walker-v0 $\leftrightarrow$ BridgeWalker-v0; PlatformJumper-v0 $\rightarrow$ GapJumper-v0), and reverse scenarios (e.g., UpStepper-v0 $\leftrightarrow$ DownStepper-v0).}
\resizebox{\columnwidth}{!}{%
\begin{tabular}{cccc}
\toprule
\textbf{Index} & \textbf{Pre-training task(s)} & \textbf{New task} & \textbf{Avg. edit dist.}\\
\midrule
1 &
Walker-v0 {\scriptsize{(easy, loc)}}, 
Pusher-v0 {\scriptsize{(easy, man)}}, 
Carrier-v0 {\scriptsize{(easy, man)}} 
&
BridgeWalker-v0 {\scriptsize{(easy, loc)}} &
18.84 {\scriptsize{(medium)}} \\

2 &
BridgeWalker-v0 {\scriptsize{(easy, loc)}} 
&
Walker-v0 {\scriptsize{(easy, loc)}} &
17.61 {\scriptsize{(medium)}} \\

3 &
Walker-v0 {\scriptsize{(easy, loc)}}, 
Carrier-v0 {\scriptsize{(easy, man)}} 
&
Pusher-v0 {\scriptsize{(easy, man)}} &
13.24 {\scriptsize{(high)}} \\

4 &
Walker-v0 {\scriptsize{(easy, loc)}}, 
Pusher-v0 {\scriptsize{(easy, man)}} 
&
Carrier-v0 {\scriptsize{(easy, man)}} &
14.46 {\scriptsize{(high)}} \\

5 &
Walker-v0 {\scriptsize{(easy, loc)}}, 
UpStepper-v0 {\scriptsize{(medium, loc)}} 
&
DownStepper-v0 {\scriptsize{(easy, loc)}} &
19.09 {\scriptsize{(medium)}} \\

6 &
Walker-v0 {\scriptsize{(easy, loc)}}, 
DownStepper-v0 {\scriptsize{(easy, loc)}} 
&
UpStepper-v0 {\scriptsize{(medium, loc)}} &
19.74 {\scriptsize{(low)}} \\

7 &
Climber-v0 {\scriptsize{(medium, cli)}} 
&
Climber-v1 {\scriptsize{(medium, cli)}} &
15.95 {\scriptsize{(high)}} \\

8 &
Climber-v0 {\scriptsize{(medium, cli)}} 
&
Climber-v2 {\scriptsize{(hard, cli)}} &
20.37 {\scriptsize{(low)}} \\

9 &
PlatformJumper-v0 {\scriptsize{(hard, loc)}} 
&
GapJumper-v0 {\scriptsize{(hard, loc)}} &
19.29 {\scriptsize{(low)}} \\
\bottomrule
\end{tabular}%
}
\label{tab:task combinations}
\end{table}


\subsubsection{Baselines}

We adopt three families of baselines: (i) evolutionary algorithms (EAs), including Bayesian Optimization (BO) \citep{kushner1964new}, Genetic Algorithm (GA) \citep{michalewicz2013genetic} and CPPN-NEAT \citep{corucci2018evolving}, which are adapted for VSRs and yield remarkable results in previous studies \citep{bhatia2021evolution}; (ii) generative approaches, including RoboGAN \citep{hu2022modular}, MorphVAE \citep{song2024morphvae}, Conditional Diffusion Model (c-DM) \citep{xu2024dynamics} and LASeR \citep{song2025laser}, which leverage deep generative models to fit and sample from high-performing robot distributions; and (iii) a POET-style cross-task transfer baseline \citep{wang2019poet}, which introduces periodic morphology migration across training tasks into GA-based optimization and directly evaluates the final-generation morphologies on the unseen task (results averaged across three independent runs; for settings with only one pre-training task, this procedure degenerates to standard GA and is marked as ``-''). For detailed introduction to these baselines, please refer to Supplementary Material C. 

\subsubsection{Evaluation metrics}
\label{metrics}

\begin{itemize}
    \item \textbf{Zero-shot fitness:} 
    \emph{Zero-shot fitness}, denoted as $f^0$, represents the fitness an algorithm can achieve on an unseen task without access to any evaluated robot samples from that particular task. For EAs, this is measured as the maximum fitness in the first generation (with population size as 25). For generative methods, it refers to the highest fitness among 25 robots generated for the new task after being fitted on training tasks. 
    \item \textbf{Relative performance:} Let $f^*_\text{a}$ be the fitness of the best-performing robot evolved by an EA, where ``a" represents the name of the specific EA. Then \emph{relative performance} is calculated as $f^0_\text{RoboLDA}/f^*_\text{a}$, comparing RoboLDA's zero-shot performance with the \emph{optimized} results of EAs. 
\end{itemize}

As illustrated in Table \ref{table:zero-shot fitness}, in five out of nine transfer settings RoboLDA uniquely achieves the highest zero-shot fitness among all reported methods, and in one additional setting it ties for the best result (BridgeWalker-v0). In the remaining three settings, another baseline attains a strictly higher fitness (Pusher-v0 and DownStepper-v0: POET; Carrier-v0: MorphVAE). Notably, the POET-style baseline demonstrates competitive performance on higher-similarity transfers where cross-task evolutionary transfer can directly leverage shared morphological features, while RoboLDA retains clear advantages on lower-similarity settings. To examine this further, we regroup the nine settings into high-, medium-, and low-similarity categories based on the average edit distance in Table \ref{tab:task combinations}. RoboLDA's advantage is most pronounced in the low-similarity group---where transfer is most challenging---winning all three settings, suggesting that it captures general hierarchical structures beyond task-specific details. It is worth noting that the advantage of RoboLDA is especially pronounced on medium and hard tasks, indicating the superior capability of RoboLDA to identify more intricate and finer-grained morphological structures. A detailed hierarchy ablation study analyzing the contribution of each level is provided in \ref{sec:hierarchy-ablation}.

\begin{table}[htbp]
\fontsize{8pt}{8pt}\selectfont
\centering
\setlength{\tabcolsep}{5pt}
\caption{Comparison of zero-shot fitness between RoboLDA and baseline algorithms, grouped by train--test task similarity measured by average edit distance (Table \ref{tab:task combinations}). As both GA and BO use random sampling for their first generation, their results are averaged. The task names are followed by their difficulty levels. Harder tasks require more complex morphologies and thus present bigger challenges for robot design.}
\begin{tabular}{lcccccccc}
\toprule
\textbf{Task} & \rotatebox{55}{\textbf{RoboLDA}} & \rotatebox{55}{\textbf{LASeR}} & \rotatebox{55}{\textbf{MorphVAE}} & \rotatebox{55}{\textbf{c-DM}} & \rotatebox{55}{\textbf{RoboGAN}} & \rotatebox{55}{\textbf{BO/GA}}  & \rotatebox{55}{{\textbf{CPPN-NEAT}}} & \rotatebox{55}{{\textbf{POET}}} \\
\midrule
\multicolumn{9}{c}{{\textbf{High-similarity transfers}}}\\
\midrule
Pusher-v0 (easy) & 11.42 & 10.97 & 11.22 & 6.99 & 7.60 & 7.79 & 9.67 & {\textbf{12.50}} \\
Carrier-v0 (easy) & 8.47 & 7.72 & \textbf{8.61} & 5.70 & 4.70 & 5.98 & 6.58 & {8.19} \\
Climber-v1 (medium) & \textbf{5.78} & 0.72 & 3.25 & 0.26 & 0.21 & 0.54 & 0.25 & {-} \\
\midrule
\multicolumn{9}{c}{{\textbf{Medium-similarity transfers}}}\\
\midrule
Walker-v0 (easy) & \textbf{10.65} & 10.62 & 10.64 & 10.21 & 8.45 & 9.85 & 6.42 & {-} \\
BridgeWalker-v0 (easy) & {\textbf{6.574}} & {\textbf{6.574}} & 6.12 & 5.14 & 3.97 & 3.94 & {6.572} & {\textbf{6.574}} \\
DownStepper-v0 (easy) & 9.07 & 5.71 & 9.06 & 7.39 & 5.92 & 6.00 & 6.21 & {\textbf{9.08}} \\
\midrule
\multicolumn{9}{c}{{\textbf{Low-similarity transfers}}}\\
\midrule
GapJumper-v0 (hard) & \textbf{5.81} & 4.13 & 4.37 & 3.63 & 3.65 & 2.65 & 3.41 & {-} \\
UpStepper-v0 (medium) & \textbf{5.13} & 2.81 & 3.65 & 2.83 & 2.71 & 3.08 & 2.41 & {4.15} \\
Climber-v2 (hard) & \textbf{1.75} & 1.68 & 0.73 & 0.29 & 0.27 & 0.23 & 0.26 & {-} \\
\bottomrule
\end{tabular}
\label{table:zero-shot fitness}
\end{table}

\begin{figure}[h]
\centering
    \includegraphics[width=0.8\linewidth]{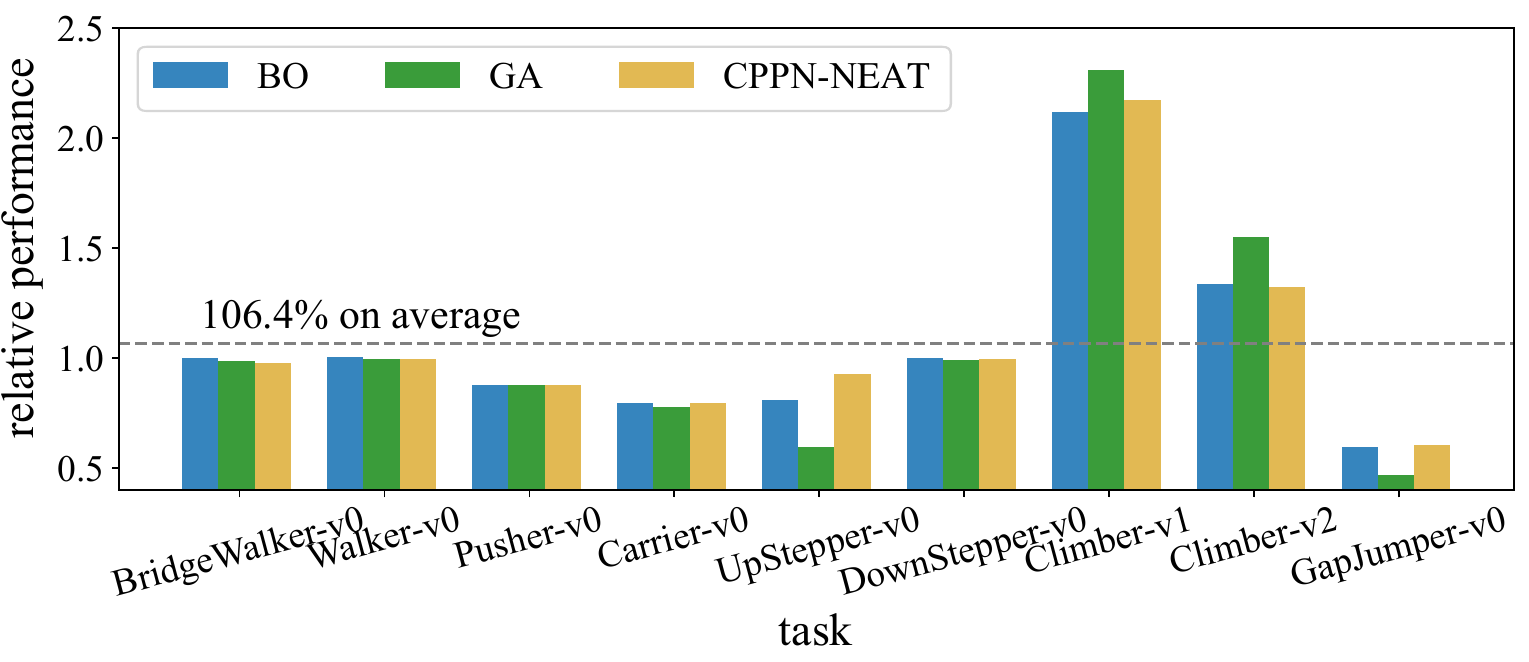}
    \caption{Relative performance of zero-shot RoboLDA compared against optimal results evolved by baseline EAs {(BO, GA, and CPPN-NEAT), using the \emph{ratio-then-average} protocol (see Section 4.4)}.}
    \label{fig:relative performance1}
\end{figure}

Now we take a bold step further and examine how the zero-shot performance of RoboLDA compares with the optimized results of EAs. Table \ref{table:generation nums} reports the numbers of generations it takes for the three EAs to reach RoboLDA's zero-shot fitness. We note that each of the EAs is allowed up to 1,000 robot evaluations to ensure thorough evolution. However, in over 55\% of cases, EAs never achieve RoboLDA's zero-shot performance (indicated as ``N/A"). It is further demonstrated in Figure \ref{fig:relative performance1} that the zero-shot-generated robots achieve 106.4\% on average of optimized fitness yielded by EAs. We note that these results are \emph{non-trivial} since RoboLDA has not seen or evaluated any robot morphology from these tasks. The advantage is especially notable in Climber-v1 and v2, which we attribute to the higher complexity of these tasks that emphasizes more informative structural priors. The above results serve as sound evidence that RoboLDA can successfully capture the generative mechanisms of existing high-performing robots and transfer the design experience to new tasks. The remarkable zero-shot performance is particularly appealing for high-stake or low-budget task settings where iterative design processes are hardly feasible. The newly generated robots already suffice for excellent performance, but can also be used as the initial population (\emph{i.e.,} an informative prior) for further evolution to yield even higher performance. 

\begin{table}[H]
\fontsize{8pt}{7pt}\selectfont
\centering
\setlength{\tabcolsep}{10pt}
\caption{The number of generations for baseline EAs to reach RoboLDA's zero-shot fitness {(using the \emph{average-then-threshold} protocol; see Section 4.4)}. ``N/A" indicates that the EA fails to do so before evolution terminates.}
\begin{tabular}{lccc}
\toprule
\textbf{Task} & \textbf{BO} & \textbf{GA} & {\textbf{CPPN-NEAT}}  \\ 
\midrule
BridgeWalker-v0 & 11 & 18 & 8 \\ 
Walker-v0 & N/A & N/A & N/A \\ 
Pusher-v0 &  {N/A} & 29 &  {N/A} \\ 
Carrier-v0 &  {N/A} & 18 &  {N/A} \\ 
UpStepper-v0 &  {N/A} & 28 &  {N/A} \\ 
DownStepper-v0 & 23 & 17 & 3 \\ 
Climber-v1 &  {N/A} &  {N/A} &  {N/A} \\ 
Climber-v2 &  {N/A} &  {N/A} &  {N/A} \\
GapJumper-v0 & 5 & 4 & 1 \\ 
\bottomrule
\end{tabular}
\label{table:generation nums}
\end{table}

\label{app:relative}


\noindent\textbf{A note on aggregation protocols.} Table \ref{table:generation nums} and Figure \ref{fig:relative performance1} employ complementary but distinct aggregation protocols because a single protocol cannot serve both purposes. In Table \ref{table:generation nums}, some EA runs never reach RoboLDA's zero-shot fitness within the evaluation budget, making the generation count N/A; a direct average across runs would be undefined. We therefore adopt an \emph{average-then-threshold} protocol: evolution curves from three runs are first averaged, and then the generation at which this averaged curve reaches RoboLDA's fitness is reported. Figure \ref{fig:relative performance1} addresses a different question---how close RoboLDA's zero-shot performance is to the EA's best-case outcome---using a \emph{ratio-then-average} protocol: relative performance (RoboLDA's zero-shot / EA's best-ever fitness) is computed per run and then averaged. Because this ratio is always defined, no N/A issue arises.

\subsection{Robot control with organ synergy}
\label{sec:control}

In this section, we explicitly showcase that the organs discerned by RoboLDA indeed bear physical significance as functional substructures. To this end, the organs are imposed as attention masks on a Transformer-based modular control policy, MetaMorph. The resulting combination, \textbf{MetaMorph-organ}, is compared with state-of-the-art synergistic control algorithms. 

\subsubsection{Baselines}

\begin{itemize}
    \item \textbf{MetaMorph} \cite{gupta2022metamorph}: 
    A Transformer-based modular control policy that decentralizes control to each actuator to learn a universal controller for different morphologies. Comparing MetaMorph-organ with MetaMorph directly reveals the effect of organ synergy. 
    \item \textbf{Synergy-oriented learning (SOLAR)} \citep{hu2022modular}: SOLAR clusters actuators into synergies according to functional similarities and basic morphological contexts. The synergies are periodically updated over control learning. It further adopts a two-level Transformer architecture that first aggregates information within each synergy and then processes information across synergies. 
    \item \textbf{SOLAR-AP}: To draw fair comparison between SOLAR and RoboLDA in respect of synergy learning, we substitute the policy network of SOLAR with MetaMorph, and apply its learned synergies as attention masks to the last self-attention layer as in MetaMorph-organ. As SOLAR employs Affinity Propagation (AP) to learn synergies, we refer to this variant as SOLAR-AP.
\end{itemize}

\subsubsection{Evaluation metric}
We use task performance as the major evaluation metric in control experiments. The task performance of each controller, represented by the discounted cumulative reward, is evaluated every few training epochs. This leads to a learning curve reflecting how well a controller is gradually learning to complete a given task. 

\begin{figure}[!htbp]
	\centering
	\subfigtopskip=0pt
	\subfigbottomskip=0pt 
	\subfigcapskip=-5pt 
        
  \subfigure[Walker-v0]{
		\label{walker-test}
		\includegraphics[width=0.4\linewidth]{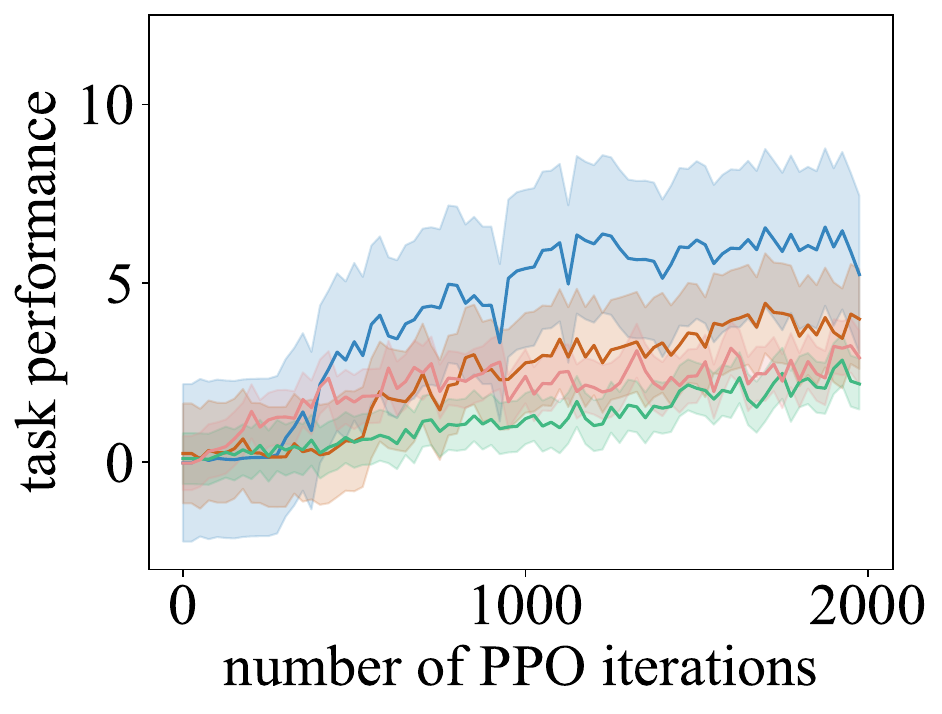}}\hspace{0pt}
\subfigure[Pusher-v0 ]{
		\label{pusher-test}
		\includegraphics[width=0.4\linewidth]{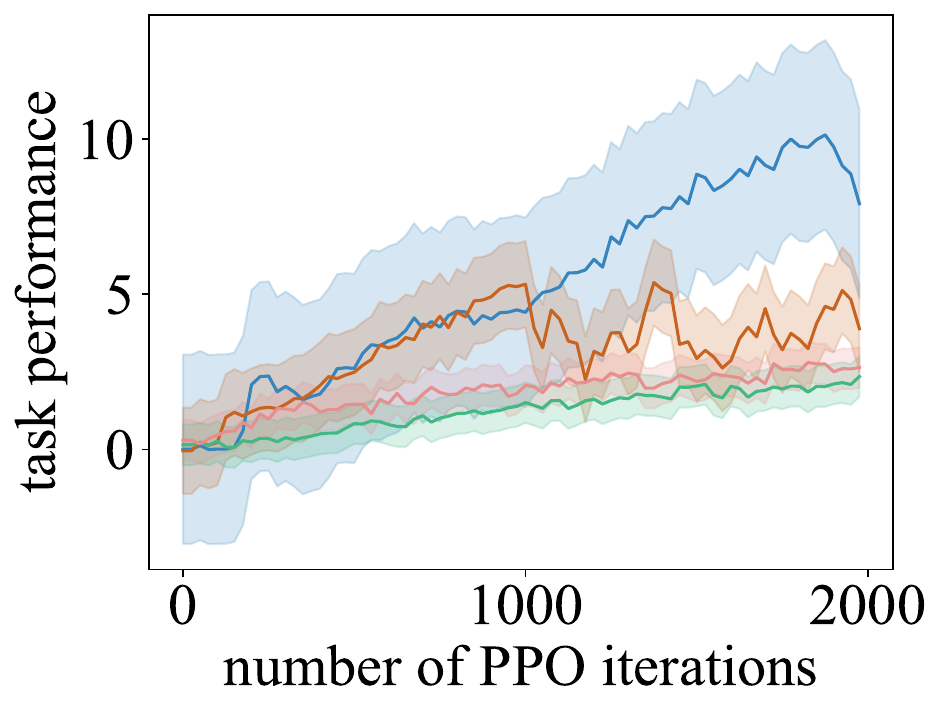}}\\
\subfigure[Carrier-v0]{
		\label{carrier-test}
		\includegraphics[width=0.4\linewidth]{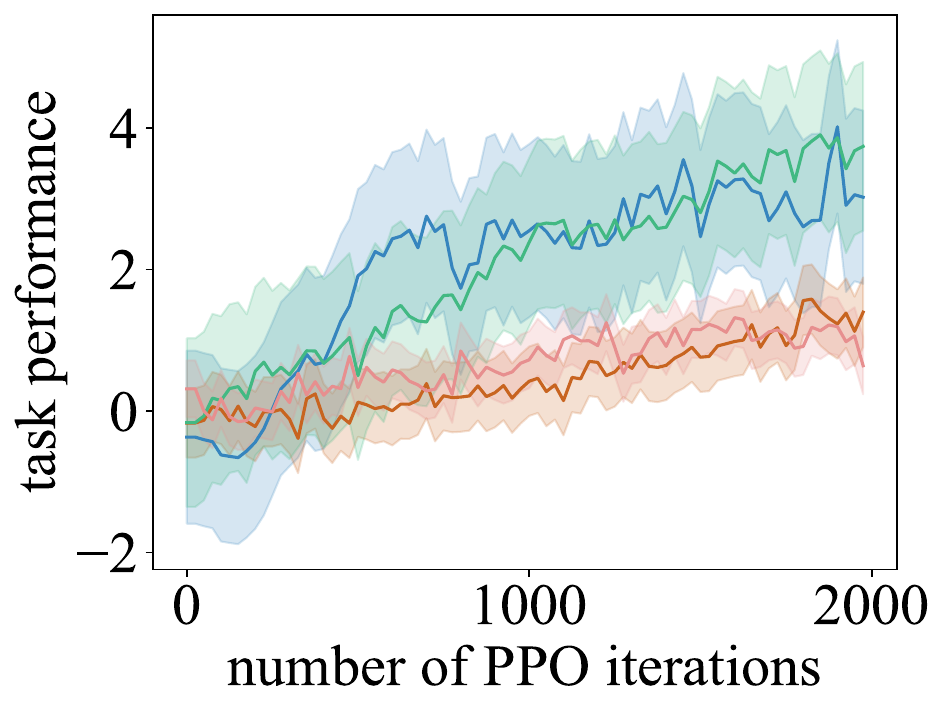}}\hspace{0pt}
\subfigure[BridgeWalker-v0]{
		\label{bridgewalker-test}
		\includegraphics[width=0.4\linewidth]{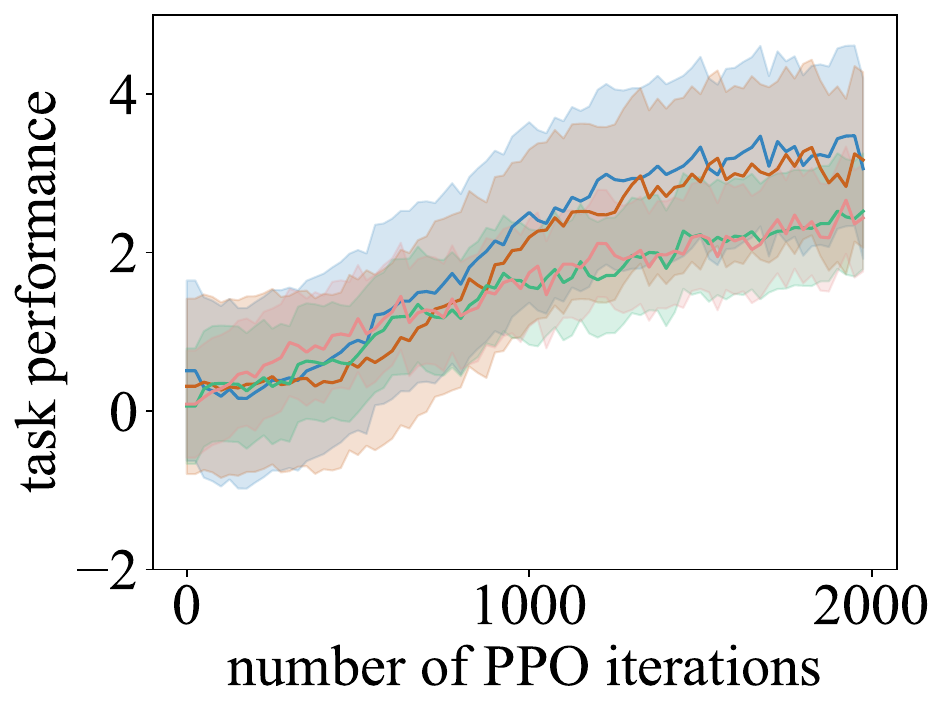}}
        \\~\\
  \subfigure{
\includegraphics[width=0.75\linewidth]{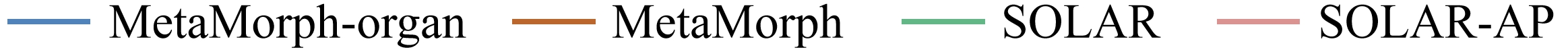}}
	\caption{Comparative results of robotic control. }
	\label{fig:control-test}
\end{figure}

\subsubsection{Comparison study}
We use two sets of task combinations, namely “Walker-v0, Pusher-v0, Carrier-v0” and “Walker-v0, BridgeWalker-v0”, in our control experiments. For each set of tasks, RoboLDA is first trained on 90\% of their high-performing robot morphologies, and then infer their organ layout according to Eq.(\ref{eq:organ}). The remaining 10\% are used to test out-of-distribution generalization, \emph{i.e.,} to examine the ability of RoboLDA to discern organ structures in unseen morphologies. In each task, for both training and test samples, 10 robots are randomly selected to undergo parallel control learning, \emph{i.e.,} with a shared policy network. 

The comparative results on test robots are presented in Figure \ref{fig:control-test} and the results on training robots are in \ref{sec:control-train}. Note that for each controller and each task, the learning curves of the ten training/test robots are averaged before being plotted. It can be seen that in almost all the cases, the organs inferred by RoboLDA (represented by MetaMorph-organ) not only facilitate generally better and faster control learning than its ablation, MetaMorph, but also outperform state-of-the-art synergy learning approaches. The superior performances on test robots additionally showcase remarkable generalizability of RoboLDA's organ inference to unseen morphologies. It is worth noting that the performance of MetaMorph-organ exhibits great robustness across different tasks, whereas the baseline algorithms all drastically fall short in some of the tasks. This suggests that decoupling synergy learning from control and delegating it to the morphological modeling of RoboLDA can facilitate more efficient and stable policy learning. Additional ablation experiments comparing RoboLDA-derived organ masks against random and adjacency-based masks are provided in \ref{sec:organ-mask-ablation}, confirming the functional specificity of the inferred organs.

We attribute such superior performances partly to improved coordination and dexterity in actuation. Please see \ref{sec:case-study} for case studies and also the code repository for qualitative results. 

\subsection{A brief analysis of computational efficiency}

It is worth noting that RoboLDA employs a probabilistic generative process for modeling robot morphology, where the internal distribution parameters are treated as \emph{latent variables} rather than trainable parameters. Specifically, the four-level hierarchy uses a single fully-connected neural network to calculate organ component weights, which involves fewer than 40k trainable parameters. Although variational inference requires an additional encoder for calculating approximate posterior, the training process is still highly efficient. With a morphology sample size of around 150 for each task, training of RoboLDA takes on average 27 seconds for each task combination, using standard laptop with consumer-level computing resources (in our case, 13th Gen Intel Core i9-13900H). 

\section{Concluding remarks}
\label{sec:conclusion}

\noindent\textbf{Conclusions.} We present RoboLDA, a pioneering hierarchical generative model for voxel-based soft robot morphology. RoboLDA assumes a four-level hierarchy encompassing ``task-robot-organ-voxel" which echoes the multilayered structures we empirically observed within high-performing VSR morphologies and is learned via a meticulously designed training procedure supported by variational inference. Through extensive experiments on simulated VSRs, we demonstrate the effectiveness of RoboLDA to infer and leverage the hierarchical priors for \emph{zero-shot} robot design, as well as the physical significance of organ structures inferred by RoboLDA to enhance synergistic motion control. 

\noindent\textbf{Limitations and future work.} Although our work mainly focuses on 2D VSRs as a proof-of-concept, which is a common benchmark used in Evolutionary Robotics, the formulation of RoboLDA is designed to be generalizable to broader settings. For instance, we have provided a straightforward extension of RoboLDA to 3D articulated rigid robots in Section \ref{sec:extension}. Extending RoboLDA to 3D VSRs (\emph{e.g.,} \citet{li2024generating}) is rather trivial: the number of voxel positions $N$ would simply become length $\times$ width $\times$ height. 

We identify several directions for future exploration. 
\textbf{First}, as an offline prior learner, RoboLDA may be less effective when optimal morphologies require fine-grained adaptations far outside the pre-training distribution, or when the collected samples are biased by premature convergence. A natural extension is to integrate RoboLDA with EDAs \citep{hauschild2011introduction} or quality-diversity methods \citep{nadizar2025enhancing,mertan2025controller} for closed-loop optimization with explicit exploration.
\textbf{Second}, although we follow the commonly adopted $5\times5$ VSR setup, RoboLDA's formulation is not tied to this resolution. Hierarchical structures are expected to be more pronounced in larger design spaces, making RoboLDA potentially more effective at higher resolutions (see \ref{sec:scalability} for a preliminary $10\times10$ experiment). 
\textbf{Finally}, all experiments are conducted in simulation, which abstracts away real-world factors such as material uncertainty and actuator nonlinearity. While sim-to-real transfer is a largely independent research challenge \citep{tobin2017domain,peng2018sim,zhao2020sim}, recent advances in pneumatic soft modules \citep{kriegman2020scalable} and reconfigurable voxel fabrication \citep{legrand2023reconfigurable} provide increasingly viable pathways. Investigating how the learned hierarchical structures transfer to physical platforms is an important future direction.



\newpage
\appendix
\section{The Derivation of Evidence Lower Bound}
\label{ELBO}

For purpose of illustration, here we present the detailed derivation of Evidence Lower Bound (ELBO) for a simplified version of RoboLDA. Specifically, we ignore the spatial positions of different voxels, but rather treat them as homogeneous. In this way, the ``global" latents $\Theta_2$ through $\Theta_4$ in the paper would all reduce to two-dimensional matrices, of sizes $H\times A$, $A\times K$ and $K\times V$, respectively, with $H$, $A$, $K$ and $V$ denoting the total numbers of task types, robot types, organ types and voxel types. The original version of RoboLDA has a much more involved ELBO that is nearly impossible to derive by hand. Therefore, we resort to the automatic calculation of {\ttfamily Pyro} (a Python package for probabilistic programming) for variational inference. 

Assume there are a total number of $D$ robot samples. Denote their morphologies (eg. voxel matrices for voxel-based soft robots) as $\boldsymbol{X}=(X_1,\cdots,X_D)$, and their task types as $\boldsymbol{h}=(h_1,\cdots,h_D)$. Note that we altered some notations compared with those in the paper to ensure a concise derivation, but the notations in this appendix are still self-consistent. 

The marginal likelihood function of observable variables, $\boldsymbol{X}$ and $\boldsymbol{h}$, is written as Eq.(\ref{eq1}), where $g_h$, $\theta_a$ and $\phi_k$ denote the parameters of multinomial distributions over robot types, organ types and voxel types, given a specific task type $h$, robot type $a$ and organ type $k$, respectively. They all follow Dirichlet prior distributions. $p(h_d)$ is the prior of task type. $Y_d$ stands for the robot type of the $d$-th robot. $Z_{di}$ and $X_{di}$ stand for the organ type and voxel type in the $i$-th voxel position, respectively. $Y_d$, $Z_{di}$ and $X_{di}$ follow multinomial distributions with $g_{h_d}$, $\theta_{Y_d}$ and $\phi_{Z_{di}}$ as parameters. Denote $(g,\theta,\phi)$ collectively as $\beta$, and denote its approximate posterior as $q(\beta|\boldsymbol{h},\boldsymbol{X})$. 

\be
\label{eq1}
\begin{aligned}
p(\boldsymbol{h},\boldsymbol{X})&=\int \prod_{h=1}^H p(g_h) \prod_{a=1}^A p(\theta_a) \prod_{k=1}^K p(\phi_k) \cdot \\
&\prod_{d=1}^D p(h_d) \left[ \int p(Y_d|g_{h_d}) \prod_{i=1}^N p(Z_{di}|\theta_{Y_d}) p(X_{di}|\phi_{Z_{di}})\d Z_d \d Y_d \right] \d g \d \theta \d \phi.
\end{aligned}
\ee

Let us denote the formula in Eq.(\ref{eq:hx}) compactly as $p(\boldsymbol{h},\boldsymbol{X}|\beta)$. According to Jensen's Inequality, we have Eq.(\ref{eq:whole}). Now we proceed to address $p(\boldsymbol{h},\boldsymbol{X}|\beta)$. Denote the approximate posterior of $Y_d$ as $q(Y_d|h_d,X_d)$. Hence, according to Jensen's Inequality, we have that Eq.(\ref{eq:yd}) holds. 

\be
\label{eq:hx}
\prod_{d=1}^D p(h_d) \left[ \int p(Y_d|g_{h_d}) \prod_{i=1}^N p(Z_{di}|\theta_{Y_d}) p(X_{di}|\phi_{Z_{di}})\d Z_d \d Y_d \right].
\ee

\be
\label{eq:whole}
\begin{aligned}
\log p(\boldsymbol{h},\boldsymbol{X})&=\log \int p(\beta) p(\boldsymbol{h},\boldsymbol{X}|\beta)\d \beta
\geq \int q(\beta|\boldsymbol{h},\boldsymbol{X}) \log \frac{p(\beta) p(\boldsymbol{h},\boldsymbol{X}|\beta)}{q(\beta|\boldsymbol{h},\boldsymbol{X})}\d \beta\\
&= \mathbb{E}_{\beta\sim q(\beta|\boldsymbol{h},\boldsymbol{X})}\log \frac{p(\beta) p(\boldsymbol{h},\boldsymbol{X}|\beta)}{q(\beta|\boldsymbol{h},\boldsymbol{X})}.
\end{aligned}
\ee

\be
\label{eq:yd}
\begin{aligned}
 &\log \int p(Y_d|h_d,g_{h_d}) \prod_{i=1}^N p(Z_{di}|\theta_{Y_d}) p(X_{di}|\phi_{Z_{di}})\d Z_d \d Y_d\\
&\geq \int q(Y_d|h_d,X_d)\log \frac{\int p(Y_d|g_{h_d})\prod_{i=1}^N p(Z_{di}|\theta_{Y_d})p(X_{di}|\phi_{Z_{di}})\d Z_d}{q(Y_d|h_d,X_d)} \d Y_d\\
&= \mathbb{E}_{Y_d\sim q(Y_d|h_d,X_d)}\log \frac{\int p(Y_d|g_{h_d})\prod_{i=1}^N p(Z_{di}|\theta_{Y_d})p(X_{di}|\phi_{Z_{di}})\d Z_d}{q(Y_d|h_d,X_d)}. 
\end{aligned}
\ee

Further denote the approximate posterior of $Z_d$ as $q(Z_d|h_d,X_d,Y_d)$. By invoking Jensen's Inequality a second time, we have Eq.(\ref{eq:zd}). Substituting Eq.(\ref{eq:zd}) into Eq.(\ref{eq:yd}), we have Eq.(\ref{eq3}). Further substituting it into Eq.(\ref{eq:hx}), we obtain Eq.(\ref{eq:hxnew}). Eventually, substitute Eq.(\ref{eq:hxnew}) into Eq.(\ref{eq:whole}) and we end up with the ELBO of the simplified version of RoboLDA, as demonstrated in Eq.(\ref{eq4}). 

\be
\label{eq:zd}
\begin{aligned}
&\log \int p(Y_d|g_{h_d})\prod_{i=1}^N p(Z_{di}|\theta_{Y_d})p(X_{di}|\phi_{Z_{di}})\d Z_d\\
&\geq \int q(Z_d|h_d,X_d,Y_d) \log \frac{p(Y_d|g_{h_d})\prod_{i=1}^N p(Z_{di}|\theta_{Y_d})p(X_{di}|\phi_{Z_{di}})}{q(Z_d|h_d,X_d,Y_d)} \d Z_d\\
&= \mathbb{E}_{Z_d\sim q(Z_d|h_d,X_d,Y_d)}\log \frac{p(Y_d|g_{h_d})\prod_{i=1}^N p(Z_{di}|\theta_{Y_d})p(X_{di}|\phi_{Z_{di}})}{q(Z_d|h_d,X_d,Y_d)}.
\end{aligned}
\ee

\be
\label{eq3}
\begin{aligned}
&\log\int p(Y_d|h_d,g_{h_d}) \prod_{i=1}^N p(Z_{di}|\theta_{Y_d}) p(X_{di}|\phi_{Z_{di}})\d Z_d \d Y_d \\
&\geq \mathbb{E}_{\substack{Y_d\sim q(Y_d|h_d,X_d)\\ Z_d\sim q(Z_d|h_d,X_d,Y_d)}} \log \frac{p(Y_d|g_{h_d})\prod_{i=1}^N p(Z_{di}|\theta_{Y_d})p(X_{di}|\phi_{Z_{di}})}{q(Y_d|h_d,X_d)q(Z_d|h_d,X_d,Y_d)}. 
\end{aligned}
\ee

\be
\label{eq:hxnew}
\begin{aligned}
\log p(\boldsymbol{h},\boldsymbol{X}|\beta)&= \sum_{d=1}^D \left[ \log p(h_d)+ \log \int p(Y_d|h_d,g_{h_d}) \prod_{i=1}^N p(Z_{di}|\theta_{Y_d}) p(X_{di}|\phi_{Z_{di}})\d Z_d \d Y_d \right]\\
& \geq \sum_{d=1}^D \left[ \log p(h_d)+ \mathbb{E}_{\substack{Y_d\sim q(Y_d|h_d,X_d)\\ Z_d\sim q(Z_d|h_d,X_d,Y_d)}} \log \frac{p(Y_d|g_{h_d})\prod_{i=1}^N p(Z_{di}|\theta_{Y_d})p(X_{di}|\phi_{Z_{di}})}{q(Y_d|h_d,X_d)q(Z_d|h_d,X_d,Y_d)}\right]. 
\end{aligned}
\ee

\be
\label{eq4}
\begin{aligned}
\log p(\boldsymbol{h},\boldsymbol{X})&\geq \sum_{d=1}^D \log p(h_d)+\mathbb{E}_{\beta\sim q(\beta|\boldsymbol{h},\boldsymbol{X})} \left[ \log p(\beta)-\log q(\beta|\boldsymbol{h},\boldsymbol{X})\right]\\
&\quad+ \sum_{d=1}^D \left[ \mathbb{E}_{\substack{\beta\sim q(\beta|\boldsymbol{h},\boldsymbol{X}) \\ Y_d\sim q(Y_d|h_d,X_d)\\ Z_d\sim q(Z_d|h_d,X_d,Y_d)}} \log \frac{p(Y_d|g_{h_d})\prod_{d=1}^N p(Z_{di}|\theta_{Y_d})p(X_{di}|\phi_{Z_{di}})}{q(Y_d|h_d,X_d)q(Z_d|h_d,X_d,Y_d)}\right]. 
\end{aligned}
\ee

\section{Pseudo Code of MetaMorph-Organ}
\label{metamorph-alg}

\begin{algorithm}[]
\fontsize{8pt}{9pt}\selectfont
\caption{Training Process of MetaMorph with Organ}
\label{alg}
\begin{algorithmic}
    \STATE {\bfseries Input:} a robot $X$ with its organs $Z$; a specific task $h$    
    \STATE {\bfseries Output:} the optimal control policy $\pi^*$
    \STATE Randomly initialize a MetaMorph policy network $\pi_\psi$    
    \STATE Initialize best task performance tracker $R^*$ as -Inf   
    \STATE Convert $Z$ into attention mask $M$ according to Eq.(6) in main text    
    \FOR{$episode = 0, 1, \ldots,n-1$} 
        \STATE trajectory = \{\}
        \STATE Reset environment of task $h$ and obtain initial state $s_0$          
        \FOR{$t = 0, 1, \ldots, T-1$}
            \STATE Execute policy $a_t  \leftarrow  \pi_{\psi}(s_t;M)$, with $M$ imposed on the last self-attention layer
            \STATE Take action $a_t$ in task environment
            \STATE Obtain reward $r_t$ and the next state $s_{t+1}$     
            \STATE trajectory = trajectory $\bigcup (s_t,a_t,r_t)$
        \ENDFOR
        \STATE Calculate surrogate loss $J_\text{PPO}(\psi)$ with the collected trajectory and update $\psi$ through multiple steps of gradient descent
        \STATE Rollout the updated policy network $\pi_{\psi}$ for a complete episode and evaluate its task performance $R$
        \IF{$R > R^*$}
            \STATE $R^* \leftarrow R$. 
            $\pi^*\leftarrow \pi_{\psi}$
        \ENDIF       
    \ENDFOR
    \STATE Return $\pi^*$
\end{algorithmic}
\end{algorithm}

We employ the Proximal Policy Optimization (PPO) \citep{schulman2017proximal} algorithm for control learning, where the surrogate loss for policy update is defined as
\begin{equation}
J_\text{PPO}(\psi)=\hat{\mathbb{E}_t}\left[  
\min\left(r_t(\psi) \hat{A}_t, \text{clip}(r_t(\psi),1-\epsilon, 1+\epsilon)\hat{A}_t\right)\right],
\end{equation}
where $\psi$ denotes trainable parameters in MetaMorph, $t$ denotes a particular time step within a trajectory,  $r_t(\psi)=\pi_\psi(a_t|s_t)/\pi_{\psi_\text{old}}(a_t|s_t)$ results from importance sampling to account for the distribution shift between $\pi_\psi$, the policy being updated, and $\pi_{\psi_\text{old}}$, the policy used for sampling trajectories. $\hat{A}_t$ is the advantage of an action that is commonly computed using the Generalized Advantage Estimator. ``clip" denotes the clip function that constrains policy updates from being too large so as to ensure training stability. Note that the training details in Algorithm \ref{alg} are simplified for brevity. In practice we utilize multiple parallel sampling processes in each iteration to improve efficiency. 

\section{Control experiments on training robots}
\label{sec:control-train}
\begin{figure*}[h]
	\centering
	\subfigtopskip=0pt
	\subfigbottomskip=0pt 
	\subfigcapskip=-5pt 
\hspace{-2mm}
  \subfigure[Walker-v0]{
		\label{walker-train}
		\includegraphics[width=0.4\linewidth]{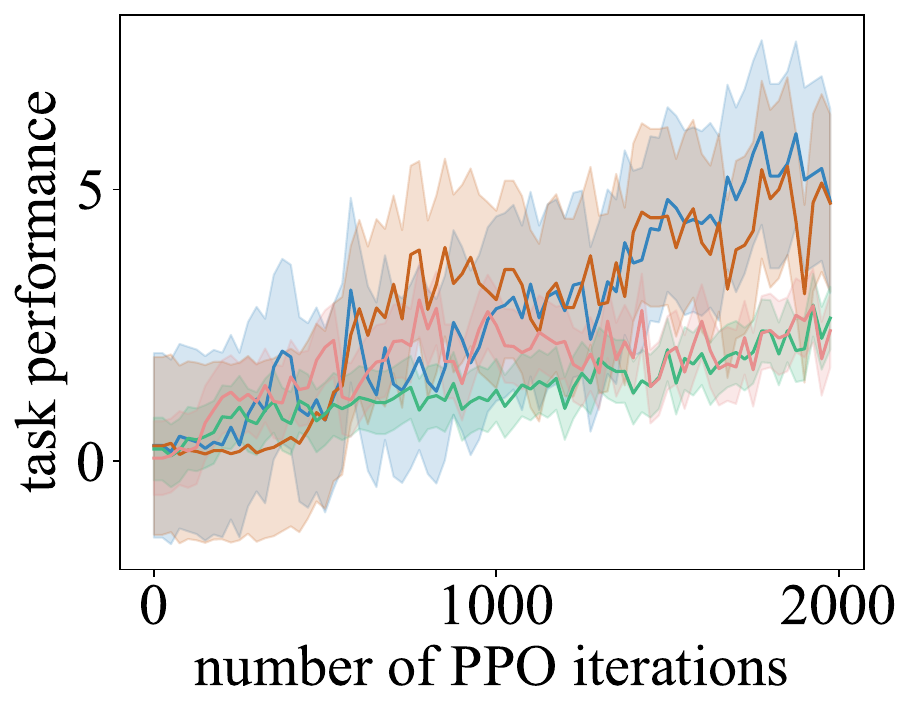}}
        \subfigure[Pusher-v0]{
		\label{pusher-train}
		\includegraphics[width=0.4\linewidth]{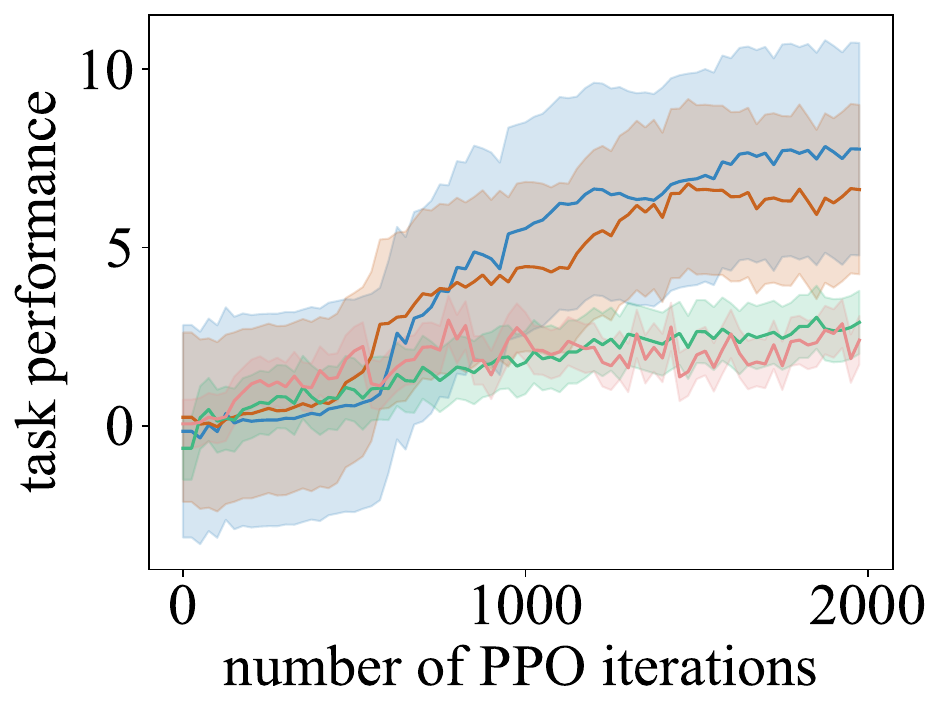}}\\
        \subfigure[Carrier-v0]{
		\label{carrier-train}
		\includegraphics[width=0.4\linewidth]{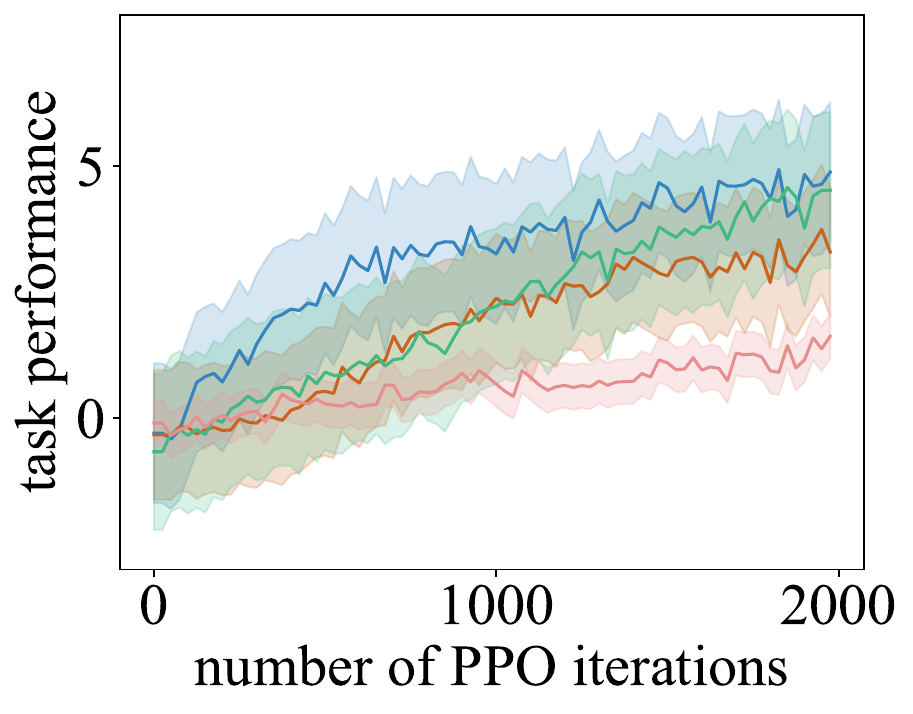}}
        \subfigure[BridgeWalker-v0]{
		\label{bridgewalker-train}
		\includegraphics[width=0.4\linewidth]{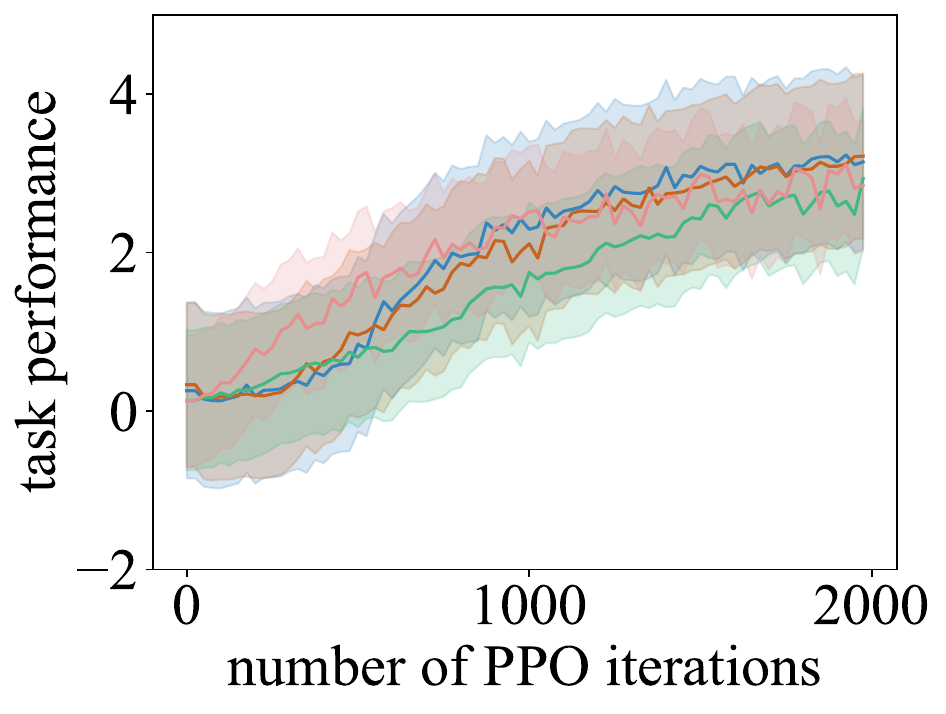}}\\~\\
  \subfigure{
\includegraphics[width=0.75\linewidth]{Figure_C.9_legend.png}}
	\caption{Results of comparison study on robotic control (training robots). }
	\label{fig:control-train}
\end{figure*}

The control learning curves of training robots are presented in Figure \ref{fig:control-train}. As with test robots, the organ structures introduce notable robustness into policy learning, yielding leading performances across different tasks. 

\section{Sensitivity analysis}
\label{sec:sensitivity}
\subsection{Hyperparameter settings}
\label{sec:hyper-sensitivity}
The numbers of robot types and organ types in RoboLDA are determined via grid search. Here we present the results and examine the sensitivity of RoboLDA to varied hyperparameter settings. Each of the hyperparameters is exhaustively searched between 4 and 8, resulting in a total of $5\times5=25$ sets of experiments. As showcased in Table \ref{table:sensitivity analysis for zero-shot robot design}, the zero-shot fitness is rather robust, indicated by coefficients of variation (COV) below 0.25 across all tasks. Tasks such as Walker-v0, BridgeWalker-v0, DownStepper-v0 and Pusher-v0 have a COV less than 0.1, demonstrating even stronger robustness, which could be attributed to their lower difficulty levels.

\begin{table*}[!h]
\fontsize{8pt}{7pt}\selectfont
\centering
\caption{Sensitivity analysis of zero-shot fitness. ``Std" and ``COV" represent standard deviation and coefficient of variation, respectively. BW, US, DS and GJ stand for BridgeWalker, UpStepper, DownStepper and GapJumper, respectively. }
\resizebox{\textwidth}{!}{%
\begin{tabular}{lccccccccc}
\toprule
\textbf{Task} & BW-v0 & Walker-v0 & Pusher-v0 & Carrier-v0 & US-v0 & DS-v0 & Climber-v1 & Climber-v2 & GJ-v0 \\ 
\midrule
\multicolumn{10}{c}{Sensitivity regarding hyperparameter settings} \\
\midrule
\textbf{Std} & 0.222 & 0.008 & 0.990 & 1.246 & 0.983 & 0.641 & 0.831 & 0.313 & 0.776 \\
\textbf{COV} & 0.034 & $<$0.001 & 0.082 & 0.141 & 0.236 & 0.072 & 0.142 & 0.159 & 0.178 \\
\midrule
\multicolumn{10}{c}{Sensitivity regarding random initialization} \\
\midrule
\textbf{Std}  & 0.004           & 0.003     & 1.213     & 1.098      & 0.006          & 0.277        & 0.050      & 0.027     & 0.375 \\
\textbf{COV}  & $<$0.001        & $<$0.001  & 0.106     & 0.130      & $<$0.001       & 0.054        & 0.009      & 0.015    & 0.102 \\
\bottomrule
\end{tabular}%
}
\label{table:sensitivity analysis for zero-shot robot design}
\end{table*}

\subsection{Random initialization}
We additionally conduct sensitivity analysis with regard to random parameter initialization. The results are obtained from three random trials. As seen in Table \ref{table:sensitivity analysis for zero-shot robot design}, the variation in Pusher-v0 and Carrier-v0 is higher than others, probably due to the additional stochasticity introduced by object manipulation. However, their COV is only slightly over 0.1, highlighting RoboLDA's stability across different parameter initializations. We believe this robustness is a unique advantage of our probabilistic approach that models distribution parameters as latent variables rather than trainable parameters.  

\section{Case studies of control experiments}
\label{sec:case-study}
\begin{figure}[h!]
    \centering
    \includegraphics[width=0.9\linewidth]{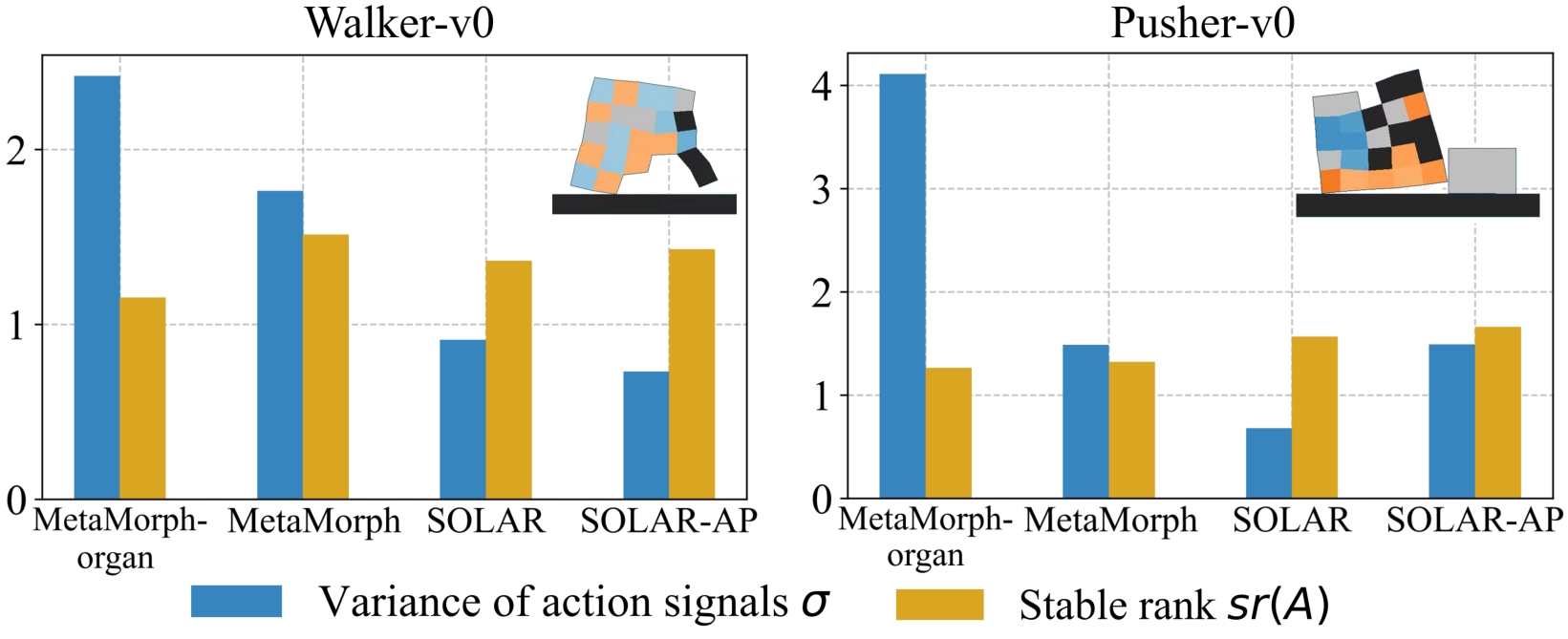}
   \caption{Case studies on coordination and dexterity. }
    \label{fig:case}
\end{figure}

We adopt the following evaluation metrics for the case studies. 
\begin{itemize}
    \item \textbf{Stable rank}: 
    As dimensionality reduction is generally accepted as an indicator of synergistic control \cite{todorov2004analysis}, we calculate the stable rank of action matrix $A$ as $sr(A)=\frac{||A||_F^2}{||A||_2^2}=\frac{\sum \sigma_i^2}{\sigma_{max}^2}$, where $A_{it}$ corresponds to the action of actuator $i$ in timestep $t$, and $\sigma_i$'s are the singular values of $A$. A smaller stable rank implies more coordinated motion across actuators. 
    \item \textbf{Variance of action signals}: 
    Define the variance of action signals as $\sigma=\frac{1}{n\cdot T}\sum_{i=1}^n \sum_{t=1}^T (A_{it}-\bar{A}_i)^2$, where $\bar{A}_i=\sum_{t=1}^T A_{it}/T$, and $n$ and $T$ denote the numbers of actuators and timesteps, respectively. According to \cite{deimel2016novel}, a larger $\sigma$ indicates greater postural variability and higher dexterity.  
\end{itemize}
We attribute the superior performances of MetaMorph-organ partly to improved coordination and dexterity in actuation. As case studies, we take two robots from ``Walker-v0" and ``Pusher-v0" as examples, and plot the variances of action signals and stable ranks yielded by different controllers in Figure \ref{fig:case}. In both cases, MetaMorph-organ exhibits a distinct advantage in generating highly coordinated action signals (\emph{i.e.,} lower stable ranks). The increased level of synergy in turn promotes finer and more dexterous motion control, marked by consistently higher variance of action signals. This is in accordance with conclusions of previous studies like \citet{santello1998postural}, regarding the benefit of synergy-based control for intricate movements. Hence, we empirically validate that the organs discerned by RoboLDA from the morphological perspective also serve superiorly as functional substructures. See the demos in our repository for qualitative results. 

\section{Validation on rigid robots}
\label{sec:rigid-validation}
\begin{figure*}[h!]
    \centering
    \includegraphics[width=0.95\linewidth]{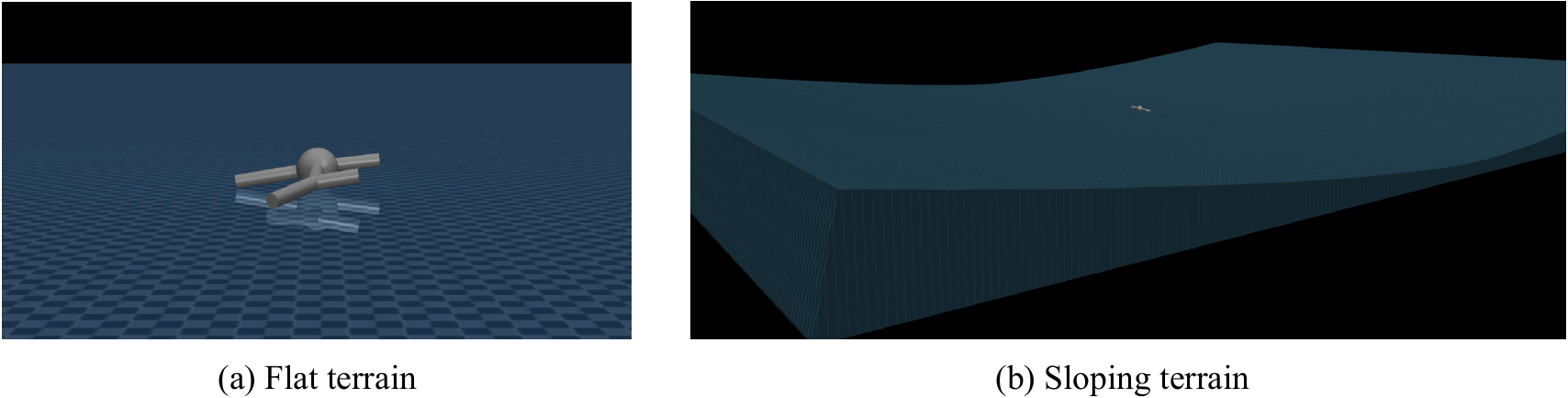}
   \caption{Locomotion environments of rigid robots. }
    \label{fig:terrains}
\end{figure*}

To validate the effectiveness of RoboLDA on rigid robots, we consider locomotion tasks on a flat terrain (training task; Figure \ref{fig:terrains}(a)) and a U-shaped sloping terrain (test task; Figure \ref{fig:terrains}(b)), with the reward function defined as the velocity along the $x$-axis. We first collect a set of high-performing robots in the training task using the genetic algorithm and then convert these robots to voxel-based representations following the procedure described in Section \ref{sec:extension}. RoboLDA is then trained on these robots and generate zero-shot designs to be evaluated in the test task. The controller of each robot design is trained with $10^5$ timesteps.

As shown in Table \ref{tab:mujoco}, the zero-shot fitness of RoboLDA outperforms baseline algorithms with a large margin. Furthermore, the relative performance of RoboLDA's zero-shot fitness with respect to GA's evolved fitness (394.46) on the sloping terrain equals \textbf{113.36\%}. The above results prove the strong applicability of RoboLDA beyond voxel-based soft robots and hopefully inspire future work to leverage hierarchical probabilistic models in wider robot design settings. The rigid robots generated by different algorithms, as well as those optimized by GA, are visualized in Supplementary Material D. 

\begin{table}[h]
\fontsize{8pt}{7pt}\selectfont
\caption{Zero-shot rigid robot design performance. }
\centering
\label{tab:mujoco}
\begin{tabular}{ccccccc}
\toprule
\textbf{Method} & GA & RoboGAN & cDM & MorphVAE & LASeR & RoboLDA \\
\midrule
\textbf{Zero-shot fitness} & 168.45 & 204.95 & 171.44 & 317.68 & 295.14 & \textbf{447.17}\\
\bottomrule
\end{tabular}
\end{table}

\section{Data assumption and limitation}

We would like to clarify that RoboLDA is not intended as a morphology optimization algorithm, but rather as a generative modeling and abstraction framework built upon existing high-performing robot designs. The process of discovering such designs—whether via evolutionary search, human expertise, or other optimization techniques—is therefore orthogonal to the proposed method and is not counted as part of RoboLDA’s computational budget. In this work, we assume the availability of a set of successful morphologies in order to study how reusable hierarchical structures can be inferred and transferred, which reflects a common setting in data-driven representation learning.

That said, we acknowledge that obtaining high-performing samples in large and complex design spaces is itself a challenging problem, and existing optimization methods may suffer from premature convergence, local optima, and sample bias, as noted in prior studies \citep{mertan2024investigating,mertan2025evolutionary}. Learning a generative model from such biased samples may limit the diversity or optimality of the generated designs, particularly when extrapolating to unseen tasks or larger design spaces. This issue is especially relevant if the learned prior is later used to initialize or guide further optimization, in which case an overly narrow prior may hinder continued exploration. Addressing this limitation is an important direction for future work. Promising avenues include integrating RoboLDA into iterative or closed-loop design pipelines, where generated samples are continuously evaluated and used to refine the training distribution, as well as combining RoboLDA with quality-diversity, migration-based/island-model, or estimation-of-distribution algorithms to better balance structural abstraction with exploration \citep{nadizar2025enhancing,mertan2025controller,bhattacharjee2019estimation}. Incorporating additional stochasticity during prior-guided search and uncertainty-aware generative modeling to mitigate overconfident extrapolation from sub-optimal data also represents an interesting future direction.

\section{Scalability analysis}
\label{sec:scalability}

To assess the scalability of RoboLDA beyond the $5\times5$ setting, we provide both a complexity analysis and a preliminary larger-scale experiment.

\textbf{Complexity.} Let $N$ denote the total number of voxels and $V$ the number of voxel types. The dominant parameter, training, and inference complexities of RoboLDA all scale as $O(N \times V)$, growing linearly with the total number of voxels. For a 2D robot with side length $s$, $N = s^2$, so moving from $5\times5$ to $10\times10$ increases cost by approximately a factor of four. For a 3D robot, $N = s^3$. Crucially, while the underlying design space grows combinatorially with resolution ($V^N$ possible morphologies), the cost of RoboLDA grows only polynomially with the voxelized representation.

\textbf{10$\times$10 experiment.} We conducted a preliminary experiment on BridgeWalker-v0 $\rightarrow$ Walker-v0 in a $10\times10$ setting. We collected high-performing morphologies from $10\times10$ BridgeWalker-v0 via GA (three repeated runs), fit RoboLDA, and generated zero-shot morphologies for Walker-v0. As shown in Table~\ref{tab:10x10}, RoboLDA's zero-shot fitness reaches 99.3\% of the GA-optimized maximum and exceeds the GA initial generation by 34.4\%, confirming that RoboLDA effectively transfers structural priors even in a substantially larger design space.

\begin{table}[h]
\fontsize{8pt}{7pt}\selectfont
\centering
\caption{Zero-shot transfer results on $10\times10$ Walker-v0. }
\label{tab:10x10}
\begin{tabular}{lc}
\toprule
\textbf{Method} & \textbf{Max fitness} \\
\midrule
RoboLDA zero-shot & 10.333 \\
GA optimized & 10.404 \\
GA generation$_0$ & 7.688 \\
\bottomrule
\end{tabular}
\end{table}

\section{Organ mask ablation}
\label{sec:organ-mask-ablation}

\begin{figure}[h!]
\centering
\begin{minipage}[t]{0.46\linewidth}\centering
\includegraphics[width=0.78\linewidth]{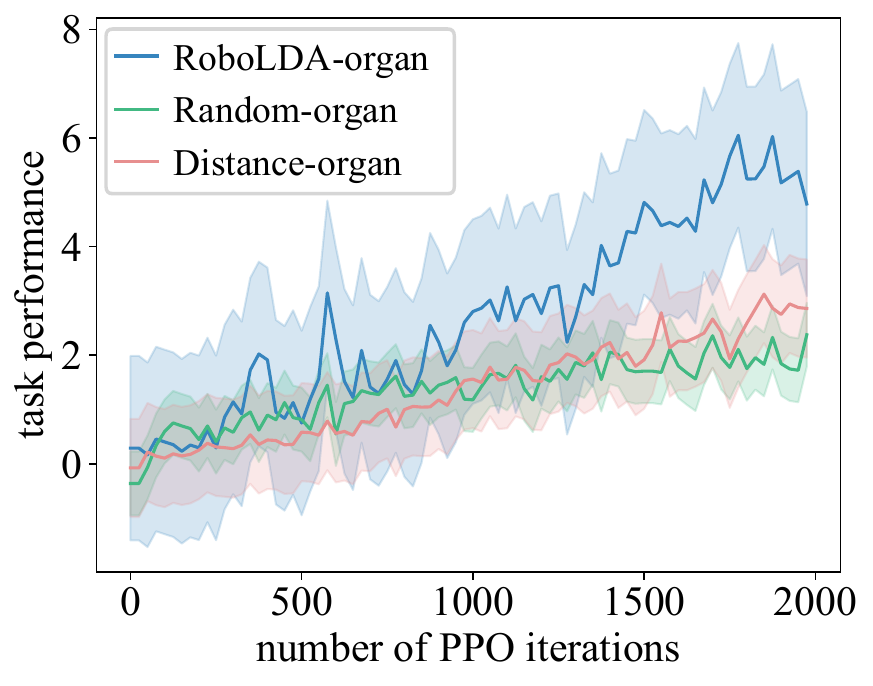}\\[-2pt]
{\small (a) Walker-v0 (train)}
\end{minipage}\hfill
\begin{minipage}[t]{0.46\linewidth}\centering
\includegraphics[width=0.82\linewidth]{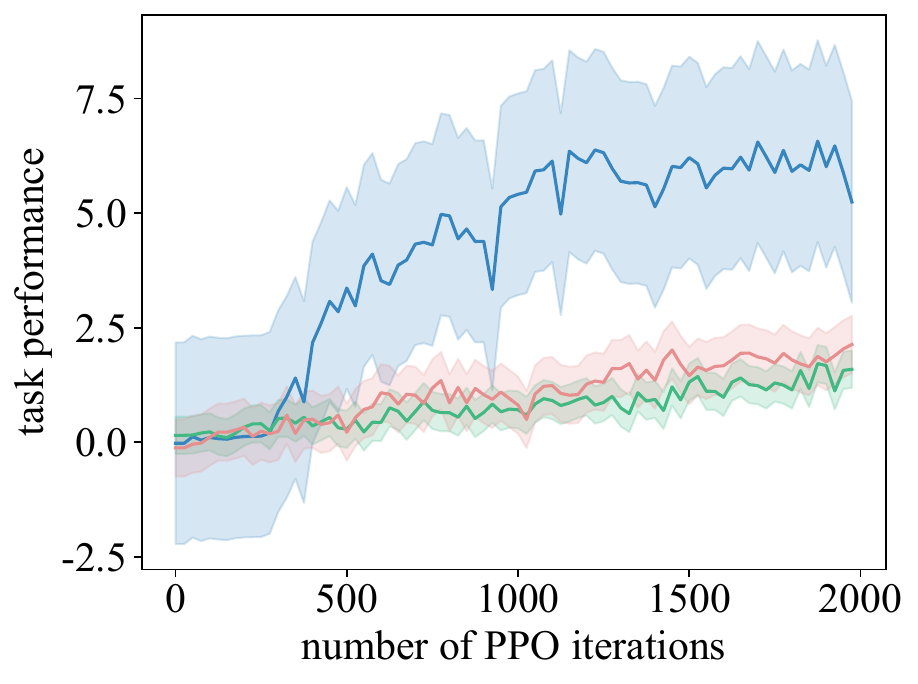}\\[-2pt]
{\small (b) Walker-v0 (test)}
\end{minipage}
\vspace{0.1em}

\begin{minipage}[t]{0.46\linewidth}\centering
\includegraphics[width=0.78\linewidth]{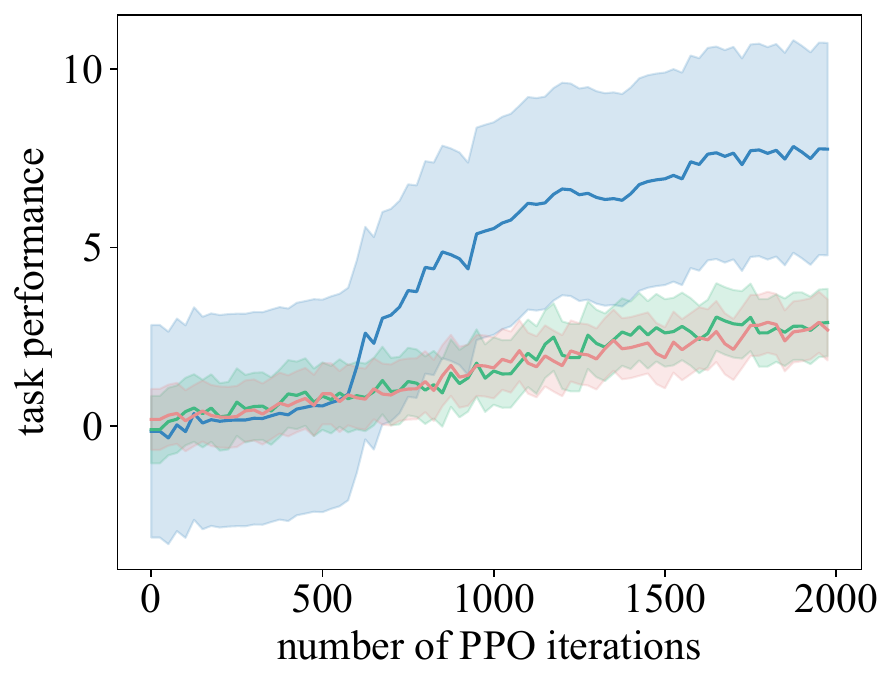}\\[-2pt]
{\small (c) Pusher-v0 (train)}
\end{minipage}\hfill
\begin{minipage}[t]{0.46\linewidth}\centering
\includegraphics[width=0.78\linewidth]{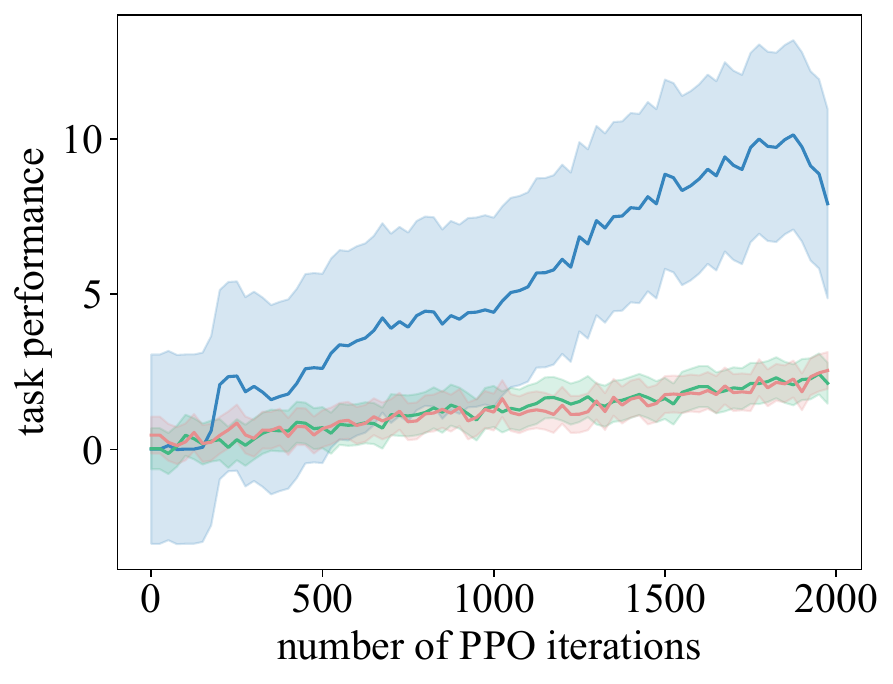}\\[-2pt]
{\small (d) Pusher-v0 (test)}
\end{minipage}
\vspace{0.1em}

\begin{minipage}[t]{0.46\linewidth}\centering
\includegraphics[width=0.78\linewidth]{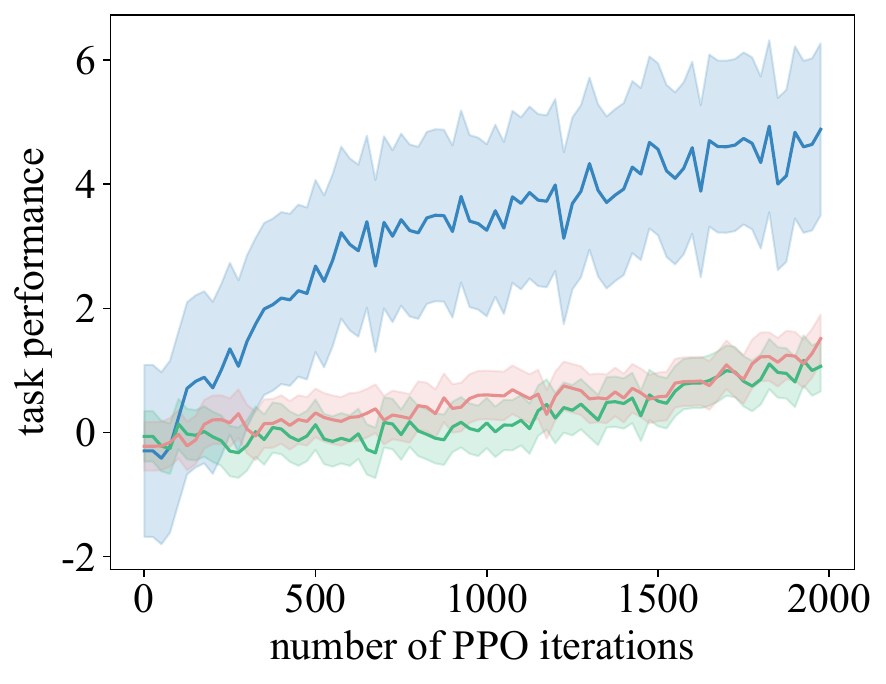}\\[-2pt]
{\small (e) Carrier-v0 (train)}
\end{minipage}\hfill
\begin{minipage}[t]{0.46\linewidth}\centering
\includegraphics[width=0.78\linewidth]{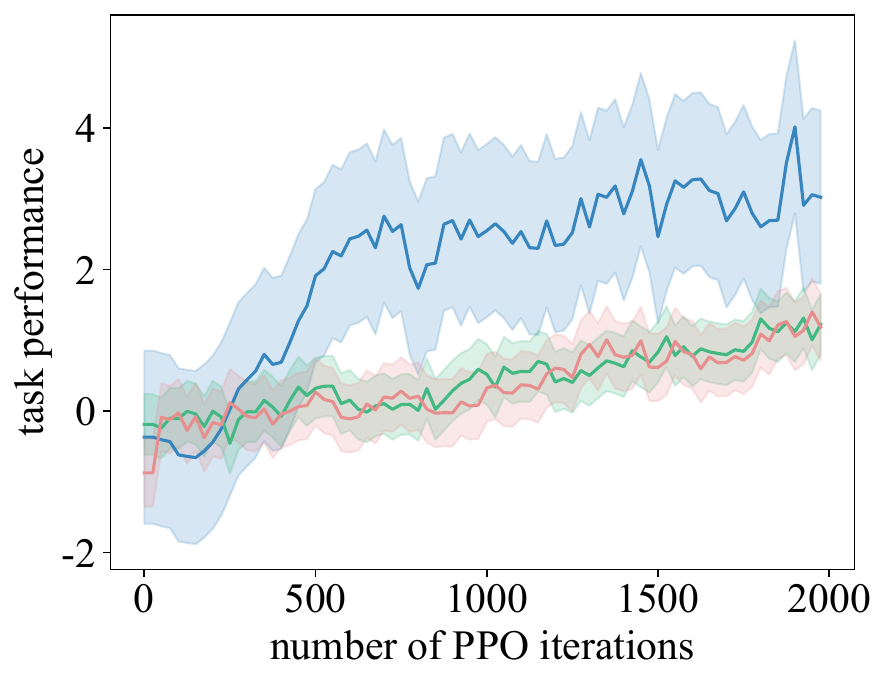}\\[-2pt]
{\small (f) Carrier-v0 (test)}
\end{minipage}
\vspace{0.1em}

\begin{minipage}[t]{0.46\linewidth}\centering
\includegraphics[width=0.78\linewidth]{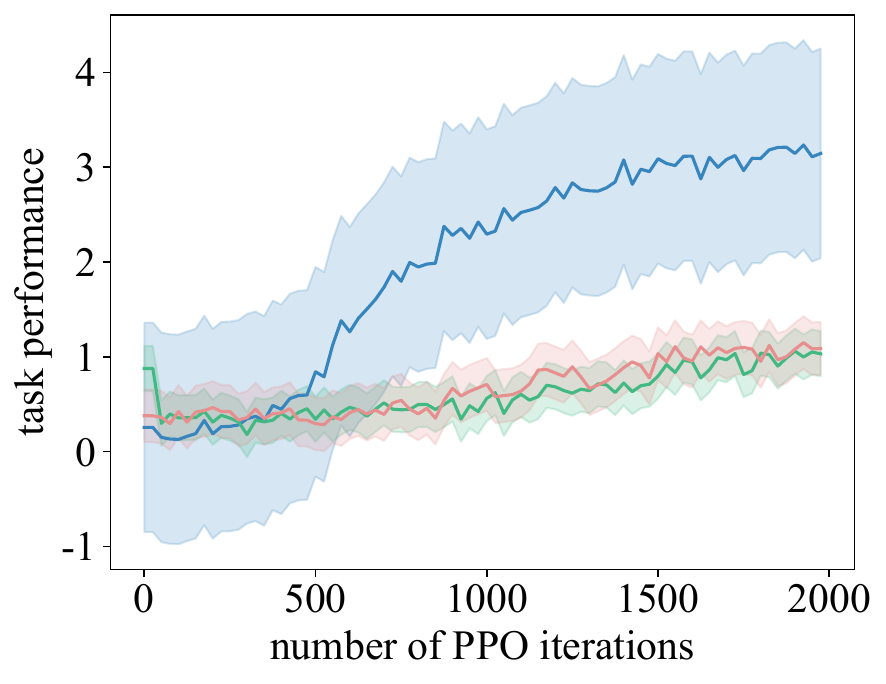}\\[-2pt]
{\small (g) BridgeWalker-v0 (train)}
\end{minipage}\hfill
\begin{minipage}[t]{0.46\linewidth}\centering
\includegraphics[width=0.78\linewidth]{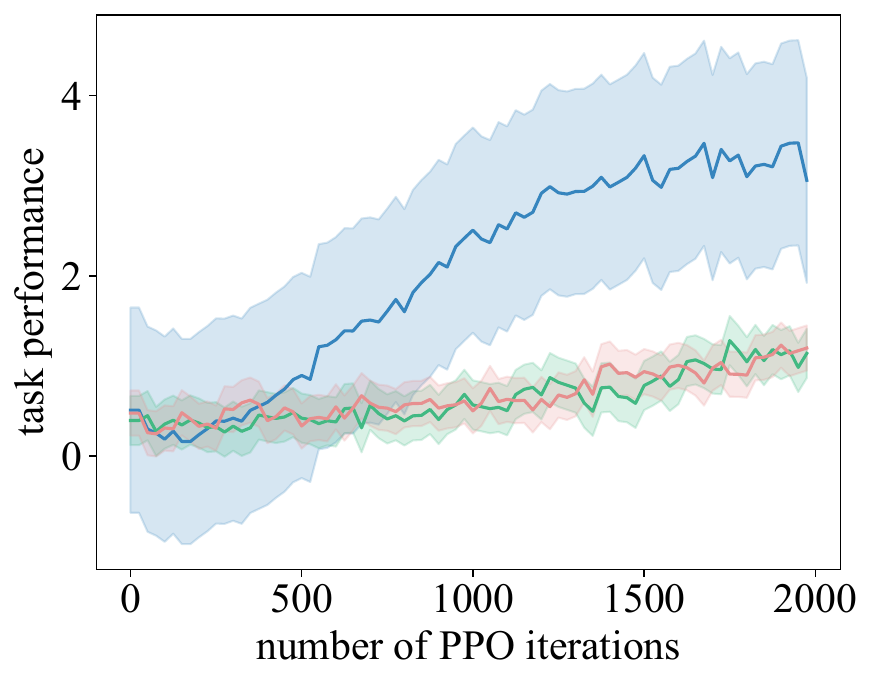}\\[-2pt]
{\small (h) BridgeWalker-v0 (test)}
\end{minipage}
\caption{Organ mask ablation on four tasks. Each row: one task; left: training split; right: held-out test split.}
\label{fig:organ-ablation}
\end{figure}

To verify that the organs inferred by RoboLDA are functionally meaningful rather than merely encoding spatial proximity, we compared three mask types in the organ-synergy control framework: (1) \textbf{RoboLDA-organ}: masks derived from RoboLDA's inferred organ layout; (2) \textbf{Random-organ}: masks generated by random grouping; and (3) \textbf{Distance-organ}: masks based on a hand-crafted adjacency prior (only spatially adjacent voxels attend to each other). As shown in Figure~\ref{fig:organ-ablation}, RoboLDA-organ consistently yields stronger learning curves on both training and test morphologies across all four tasks, supporting the claim that the inferred organs capture task-relevant functional substructures.

\section{Hierarchy ablation}
\label{sec:hierarchy-ablation}

To analyze the contribution of each level of the hierarchy, we evaluated four ablated variants on three representative transfer settings: Climber-v0 $\rightarrow$ Climber-v1 (high similarity), BridgeWalker-v0 $\rightarrow$ Walker-v0 (medium), and Walker-v0 \& DownStepper-v0 $\rightarrow$ UpStepper-v0 (low). The variants are: (1) \textbf{w/o position modeling}: all position-dependent modeling is flattened; (2) \textbf{task-organ-voxel}: the robot layer is removed; (3) \textbf{task-robot-voxel}: the organ layer is removed; and (4) \textbf{task-voxel}: both intermediate layers are collapsed. As shown in Table~\ref{tab:hierarchy-ablation}, all ablated variants perform worse than the full model, with the degradation most pronounced on low-similarity transfers, confirming that the complete hierarchy helps capture more transferable structural regularities.

\begin{table}[h]
\fontsize{8pt}{7pt}\selectfont
\centering
\caption{Hierarchy ablation on three representative transfer settings. }
\label{tab:hierarchy-ablation}
\begin{tabular}{lccc}
\toprule
\textbf{Method} & Climber-v1 (high) & Walker-v0 (med.) & UpStepper-v0 (low) \\
\midrule
RoboLDA & \textbf{5.78} & \textbf{10.65} & \textbf{5.13} \\
w/o position modeling & 4.13 & 10.59 & 4.02 \\
task-organ-voxel & 5.55 & 10.62 & 4.16 \\
task-robot-voxel & 5.72 & 10.62 & 3.93 \\
task-voxel & 5.70 & 10.63 & 4.29 \\
\bottomrule
\end{tabular}
\end{table}

\section{Extended statistical evaluation (10 runs)}
\label{sec:10run}

To provide stronger statistical support, we ran 10 independent trials on the four most challenging transfer settings, comparing RoboLDA against MorphVAE and LASeR (the two strongest generative baselines) as well as BO/GA and CPPN-NEAT. As shown in Table~\ref{tab:10run}, RoboLDA consistently achieves the highest mean fitness with moderate variance, confirming that the advantages reported in Table~\ref{table:zero-shot fitness} are robust across random seeds.

\begin{table}[h]
\fontsize{8pt}{7pt}\selectfont
\centering
\caption{Zero-shot fitness (mean $\pm$ std) over 10 independent runs on four challenging transfer settings.}
\label{tab:10run}
\begin{tabular}{lccccc}
\toprule
\textbf{Task} & \textbf{RoboLDA} & MorphVAE & LASeR & BO/GA & CPPN-NEAT \\
\midrule
UpStepper-v0 & \textbf{4.62} $\pm$ 0.71 & 3.72 $\pm$ 1.81 & 1.55 $\pm$ 0.63 & 2.86 $\pm$ 0.82 & 4.03 $\pm$ 0.56 \\
Climber-v1 & \textbf{5.92} $\pm$ 0.84 & 4.35 $\pm$ 0.71 & 0.89 $\pm$ 0.08 & 0.27 $\pm$ 0.04 & 1.88 $\pm$ 2.20 \\
Climber-v2 & \textbf{1.78} $\pm$ 0.07 & 1.11 $\pm$ 0.43 & 1.43 $\pm$ 0.39 & 0.27 $\pm$ 0.04 & 0.97 $\pm$ 0.57 \\
GapJumper-v0 & \textbf{5.78} $\pm$ 0.86 & 4.45 $\pm$ 1.10 & 4.29 $\pm$ 0.14 & 2.40 $\pm$ 0.67 & 4.47 $\pm$ 1.39 \\
\bottomrule
\end{tabular}
\end{table}


The supplementary materials are organized as follows: In Section A, we provide an introduction to task environments used in this work; In Section B, we present our hyperparameter settings to ensure reproducibility; In Section C, we provide detailed descriptions of baseline algorithms involved in robot design experiments (including the newly added POET baseline); Section D visualizes robot morphologies produced by different algorithms in rigid robot experiments; Section E includes additional visualizations of VSR hierarchical structures learned by RoboLDA. Additional analyses added during revision---including scalability analysis ($10\times10$ grid), RL algorithm comparison, organ mask ablation, hierarchy ablation, and extended statistical evaluation (10 runs)---are included as appendix sections in the main manuscript.

\section{Introduction to task environments}
\label{sec:tasks}

In this section, we present a brief introduction to the tasks we adopted for benchmarking in Evolution Gym. The introduction is heavily borrowed from their original paper \cite{bhatia2021evolution}. 

Let us first define some notations that would be used later.
\begin{itemize}
    \item \textbf{Position: } Denote with $p^o$ the position of the center of mass of an object $o$, which consists of two components $p_x^o$ and $p_y^o$, \emph{i.e.,} the positions on $x$ and $y$ axis. $p^o$ is derived by averaging the positions of all the point-masses that make up object $o$;
    \item \textbf{Velocity: }Denote with $v^o$ the velocities of the center of mass of an object $o$, which consists of two components $v_x^o$ and $v_y^o$, \emph{i.e.,} the velocity on $x$ and $y$ axis. $v^o$ is computed by averaging the velocities of all point masses that make up object $o$; 
    \item \textbf{Orientation: }Denote with $\theta^o$, a vector of length one, the orientation of an object $o$. Denote the position of point mass $i$ of object $o$ as $p_i$, and $\theta^o$ is computed by averaging over all $i$ the angle between the vector $p_i-p^o$ at current time and the initial state. This average is weighted by $||p_i-p^o||$ in the initial state. 
    \item \textbf{Other observations: }Let $c^o$ be a vector of length $2n$ that describes the relative positions of all $n$ point masses of object $o$ to the center of mass. Let $h_b^o(d)$ characterize the terrain information around a robot below its center of mass. More specifically, for some integer $x\leq d$, the corresponding entry in vector $h_b^o(d)$ will be the highest point of the terrain which is lower than $p_y^o$ between a range of $[x,x+1]$ voxels from $p_x^o$ in the $x$-direction. 
    \item Besides, we would denote the robot as object $r$, the box that it is trying to manipulate as object $b$, the number of point masses in $r$ as $n$, the observation vector as $\mathcal{S}$, and the reward function as $R$.  
\end{itemize}

\subsection{Walker-v0}

\begin{figure}[htbp] 
	\centering  
	\includegraphics[width=\linewidth]{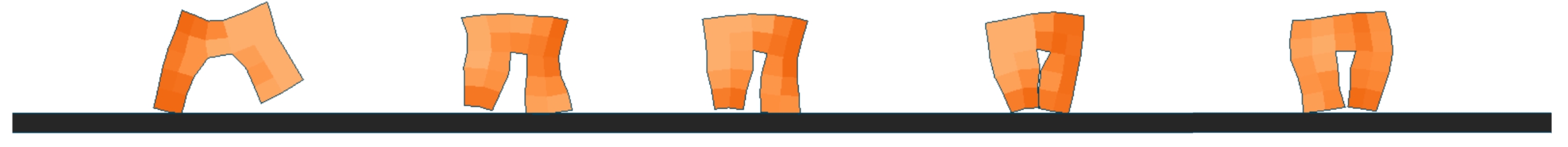}
	  
 \caption{Walker-v0}
	\label{fig:walker}
\end{figure}
In this task, the robot is required to walk as far as possible on flat terrain. $\mathcal{S}\in \mathbb{R}^{n+2}$ consists of $v^r$ and $c^r$ with lengths 2 and $n$. $R=\Delta p_x^r$ rewards the robot for moving in the positive $x$-direction. The robot is also given a one-time reward of $1$ for reaching the end of the terrain. 

\subsection{Pusher-v0}

\begin{figure}[htbp] 
	\centering  
	\includegraphics[width=\linewidth]{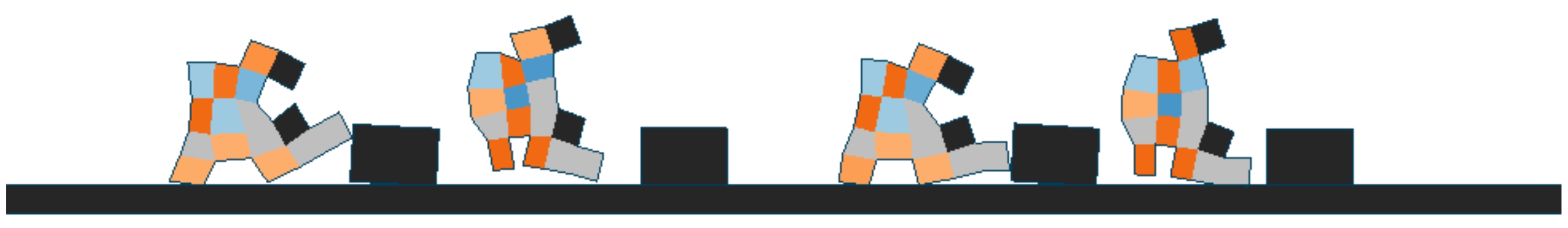}
 \caption{Pusher-v0}
	\label{fig:pusher}
\end{figure}
In this task, the robot is required to push a box initialized in front of it. $\mathcal{S}\in \mathbb{R}^{n+6}$ consists of $v^b$, $p^b-p^r$, $v_r$ and $c^r$ with lengths 2, $n$, 2 and 2 respectively. $R=R_1+R_2$, where $R_1=0.5\cdot \Delta p_x^r + 0.75 \cdot \Delta p_x^b$ rewards the robot and the box for moving in the positive $x$-direction, and $R_2=-\Delta |p_x^b - p_x^r|$ penalizes the robot and the box for separating in the $x$-direction. The robot is also given a one-time reward of $1$ for reaching the end of the terrain. 

\subsection{Carrier-v0}
\begin{figure}[htbp] 
	\centering  
	\includegraphics[width=\linewidth]{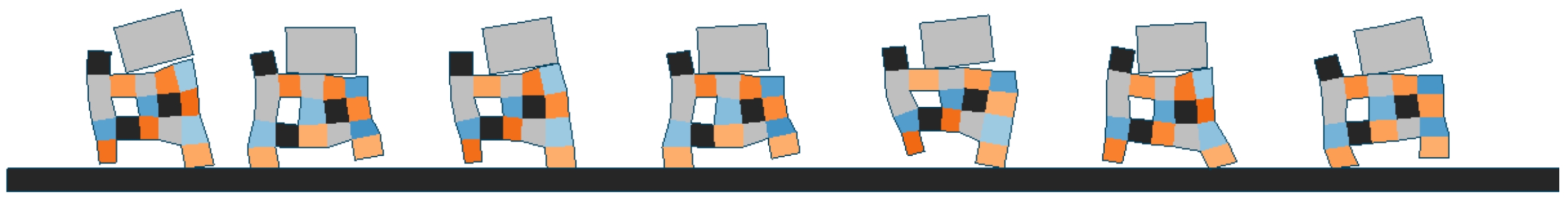}
	  
 \caption{Carrier-v0}
	\label{fig:carrier}
\end{figure}
In this task, the robot is required to catch a box initialized above it and carries it as far as possible. $\mathcal{S}\in \mathbb{R}^{n+6}$ consists of $v^b$, $p^b-p^r$, $v^r$ and $c^r$ with lengths 2, $n$, 2 and 2 respectively. $R=R_1+R_2$, where $R_1=0.5\cdot \Delta p_x^r+0.5\cdot \Delta p_x^b$ rewards the robot and the box for moving in the positive $x$-direction, and $R_2=0$ if $p_y^b\geq t_y$ and otherwise $10\cdot \Delta p_y^b$ penalizes the robot for dropping the box below a threshold height $t_y$. The robot is also given a one-time reward of $1$ for reaching the end of the terrain. 

\subsection{BridgeWalker-v0}
\begin{figure}[H] 
	\centering  
	\includegraphics[width=\linewidth]{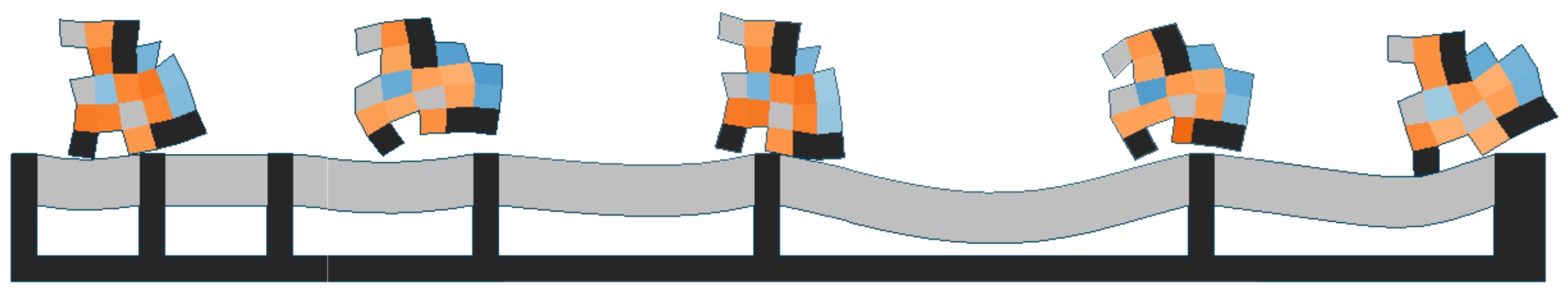}
	  
 \caption{BridgeWalker-v0}
	\label{fig:bridgewalker}
\end{figure}
In this task, the robot is required to walk as far as possible on a soft rope-bridge. $\mathcal{S}\in \mathbb{R}^{n+3}$ consists of $v^r$, $\theta^r$ and $c^r$ with lengths 2, 1 and $n$ respectively. $R=\Delta p_x^r$ rewards the robot for moving in the positive $x$-direction. The robot is also given a one-time reward of 1 for reaching the end of the terrain. 

\subsection{UpStepper-v0}
\begin{figure}[H] 
	\centering  
	\includegraphics[width=\linewidth]{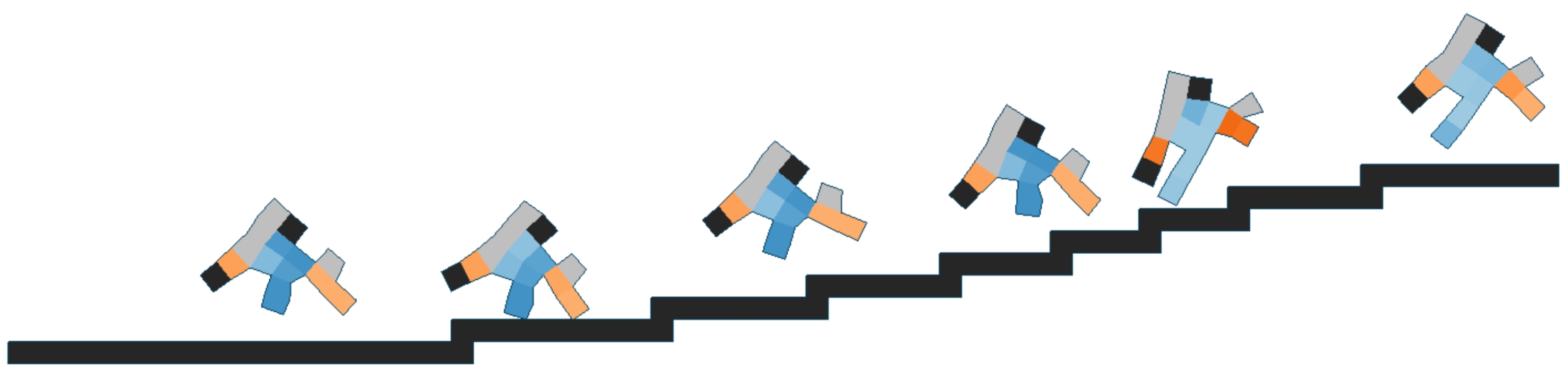}
	  
 \caption{UpStepper-v0}
	\label{fig:upstepper}
\end{figure}
In this task, the robot is required to mount stairs of varying lengths. $\mathcal{S}\in\mathbb{R}^{n+14}$ consists of $v^r$, $\theta^r$, $c^r$ and $h_b^r(5)$ with lengths 2, 1, $n$ and 11, respectively. $R=\Delta p_x^r$ rewards the robot for moving in the positive $x$-direction. The robot is given a one-time reward of 2 for reaching the end of the terrain, and a one-time penalty of -3 for rotating more than 75 degrees from its original orientation in either direction (after which the environment is reset). 

\subsection{DownStepper-v0}
\begin{figure}[htbp] 
	\centering  
	\includegraphics[width=\linewidth]{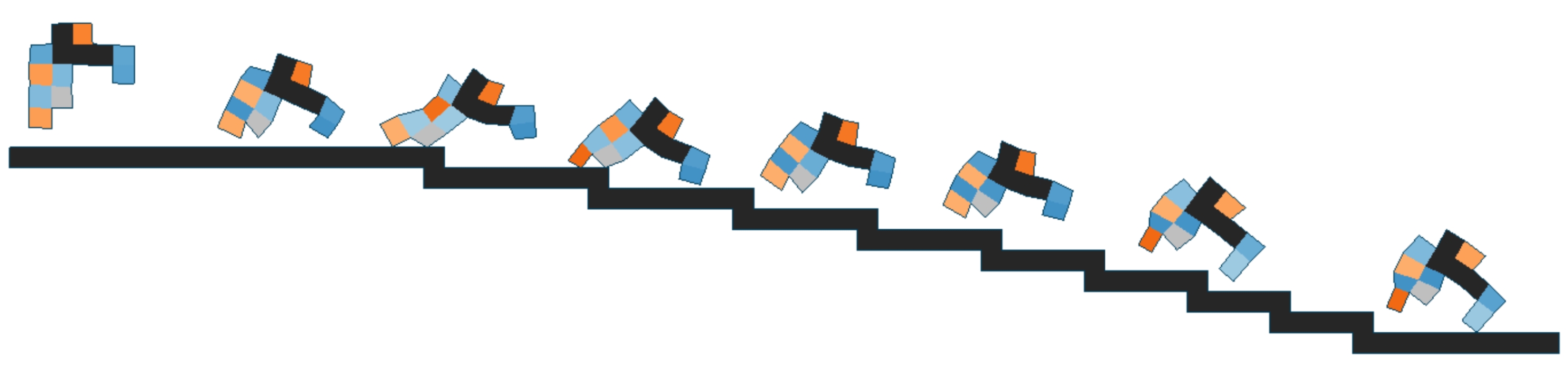}
	  
 \caption{DownStepper-v0}
	\label{fig:downstepper}
\end{figure}
In this task, the robot is required to climb down stairs of varying lengths. $\mathcal{S}\in\mathbb{R}^{n+14}$ consists of $v^r$, $\theta^r$, $c^r$ and $h_b^r(5)$ with lengths 2, 1, $n$ and 11 respectively. $R=\Delta p_x^r$ rewards the robot for moving in the positive $x$-direction. The robot is given a one-time reward of 2 for reaching the end of the terrain, and a one-time penalty of -3 for rotating more than 90 degrees from its original orientation in either direction (after which the environment is reset). 

\subsection{Climber-v0}
\begin{figure}[H] 
	\centering  
    \rotatebox{-90}{\includegraphics[width=0.2\linewidth]{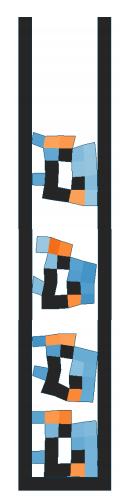}}
	  
 \caption{Climber-v0. The figure has been rotated 90 degrees clockwise to save space, where the vertical line on the left represents the ground. The same is true for Figures \ref{fig:climber1} and \ref{fig:climber2}. }
	\label{fig:climber0}
\end{figure}

In this task the robot climbs as high as possible through a flat, vertical channel. $\mathcal{S}\in\mathbb{R}^{n+2}$, consists of $v^r$ and $c^r$ with lengths 2, and $n$ respectively.$R=\Delta p_y^r$ rewards the robot for moving in the positive $y$-direction. The robot is given a one-time reward of 1 for reaching the end of the terrain(after which the environment is reset). 

\subsection{Climber-v1}

\begin{figure}[H] 
	\centering  
    \rotatebox{-90}{\includegraphics[width=0.13\linewidth]{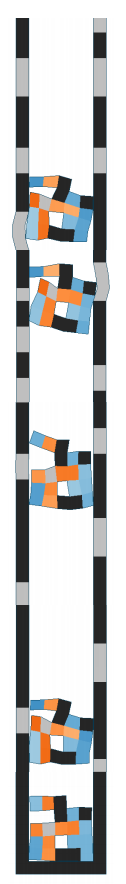}}
	  
    \caption{Climber-v1}
	\label{fig:climber1}
\end{figure}

In this task the robot climbs as high as possible through a vertical channel made of mixed rigid and
soft materials. $\mathcal{S}\in\mathbb{R}^{n+2}$, consists of $v^r$ and $c^r$ with lengths 2, and $n$ respectively.$R=\Delta p_y^r$ rewards the robot for moving in the positive $y$-direction. The robot is given a one-time reward of 1 for reaching the end of the terrain(after which the environment is reset). 

\subsection{Climber-v2}
\begin{figure}[htbp] 
	\centering  
	\rotatebox{-90}{\includegraphics[width=0.2\linewidth]{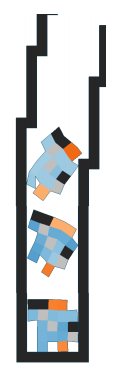}}
	  
 \caption{Climber-v2}
	\label{fig:climber2}
\end{figure}

In this task the robot climbs as high as possible through a narrow stepwise channel.  $\mathcal{S}\in\mathbb{R}^{n+10}$, consists of $v^r$, $\theta^r$, $c^r$ and $h_a^r(3)$ with lengths 2, 1, $n$ and 7 respectively. $R=\Delta p_y^r+0.2\Delta p_x^r$ rewards the robot for moving in the positive $y$-direction and positive $x$-direction. 

\subsection{PlatformJumper-v0}

\begin{figure}[htbp]
	\centering  
\includegraphics[width=\linewidth]{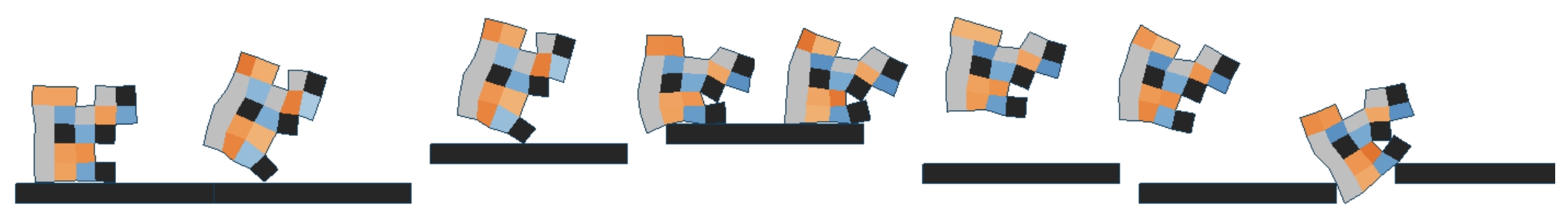}
 \caption{PlatformJumper-v0}
	\label{fig:platformjumper-v0}
\end{figure}

In this task, the robot traverses a series of floating platforms at different heights. $\mathcal{S}\in\mathbb{R}^{n+14}$ consist of $v^r, \theta^r, c^r, ~\text{and} ~h_b^r(5)$ with lengths of 2, 1, $n$, and 11, respectively. $R=\Delta p_x^r$ rewards the robot for moving in the positive $x$-direction. The robot is also given a one-time penalty of -3 for rotating more that 90 degrees from its original orientation in either direction or for falling off the platforms (after which the environment is reset). 

\subsection{GapJumper-v0}

\begin{figure}[htbp] 
	\centering  
	\includegraphics[width=\linewidth]{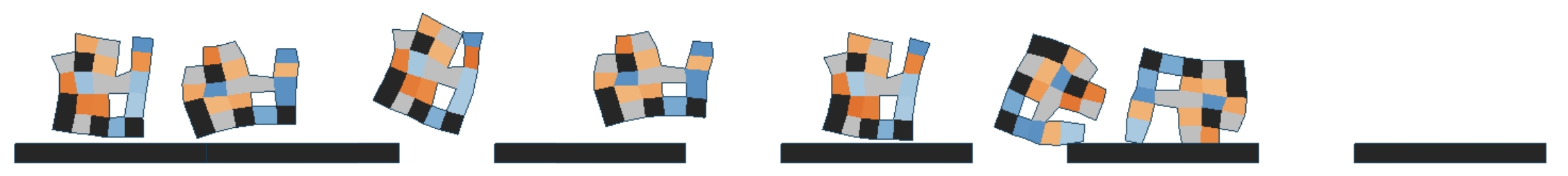}
 \caption{GapJumper-v0}
	\label{fig:gapjumper-v0}
\end{figure}

In this task, the robot traverses a series of spaced-out floating platforms all at the same height. $\mathcal{S}\in\mathbb{R}^{n+14}$ consists of $v^r, \theta^r, c^r, ~\text{and} ~h_b^r(5)$ with lengths of 2, 1, $n$, and 11, respectively. $R=\Delta p_x^r$ rewards the robot for moving in the positive $x$-direction. The robot is also given a one-time penalty of -3 for falling off the platforms (after which the environment is reset). 

\section{Hyperparameter settings}
\label{sec:hyperparameters}

For reproducibility, here we list all our hyperparameters adopted for RoboLDA, the PPO algorithm and rigid robot extension in Table \ref{tab:params}. 

\begin{table}[!ht]
\fontsize{8pt}{11pt}\selectfont
\centering
\caption{Hyperparameter settings.}
\begin{tabular}{cc}
\toprule    
\textbf{Hyperparameter} & \textbf{Value} \\ 
\midrule 
\multicolumn{2}{c}{RoboLDA} \\
\midrule 
robot size & $5\times 5$\\
number of robot types & 6 \\
number of organ types & 6 \\
number of organ components & 10 \\
hidden dims of MLPs & [128]*3 \\
activation function in MLPs & Tanh \\
learning rate & $10^{-4}$ \\
number of training epochs & 400 \\
$\beta_1$ and $\beta_2$ in ADAM optimizer & (0.95,0.999) \\
\midrule 
\multicolumn{2}{c}{PPO Policy Training} \\
\midrule 
number of parallel sampling processes & 4 \\
number of time steps in each process & 128\\
learning rate & $2.5\times 10^{-4}$ \\
$\epsilon$ in the clip function of PPO & 0.1\\
number of iterations (MLP control/modular control) & 1000/2000 \\
number of epochs per iteration & 4 \\
number of mini-batches per epoch & 4 \\
$\lambda$ in generalized advantage estimation (GAE) & 0.95 \\
\midrule 
\multicolumn{2}{c}{Rigid robot configurations} \\
\midrule 
Torso radius              & 0.2                      \\
Torso mass                & 0.5                      \\
Limb lengths              & 0.25/0.5/0.75            \\
Limb radius               & 0.1                     \\
Limb mass                 & 0.5                      \\
Joint rotation directions & $x$/$y$/$z$              \\
Joint rotation limits     & (-0.6, 0.6) (in radians)  \\
\bottomrule 
\end{tabular}
\label{tab:params}
\end{table}

\section{Detailed introduction to robot design baselines}
\begin{itemize}

\item \textbf{Bayesian Optimization} (BO; \citealp{kushner1964new}): BO is an evolutionary strategy that utilizes the principles of Bayesian theory to guide global optimization. It is particularly suitable for problems with high evaluation costs. In this work, we employ the Gaussian Process as the surrogate model and expected improvement as the acquisition function.

\item \textbf{Genetic Algorithm} (GA; \citealp{michalewicz2013genetic}): GA is a search algorithm that simulates the biological evolution process in nature. It starts with an initial population of robot designs and evolves the population through selection and random mutation, producing morphologies with higher fitness over generations. 

\item \textbf{CPPN-NEAT} \cite{corucci2018evolving}: one of the predominant methods for soft robot design automation, combining the pattern generation capabilities of Compositional Pattern Producing Networks (CPPNs) with the neural architecture evolution mechanism of NeuroEvolution of Augmenting Topologies (NEAT). A robot design is obtained by querying a CPPN with all spatial coordinates in the voxel matrix. 

\item \textbf{RoboGAN} \cite{hu2022modular}: RoboGAN employs the generative adversarial network (GAN; \citealp{goodfellow2020generative}) for the modeling of robot morphology. It treats the original robot designs generated by GAN as negative samples, which are then optimized with a few steps of evolutionary search to derive positive samples. A generator and a discriminator are updated iteratively so that the former better approximates high-performing robot distributions.  

\item \textbf{MorphVAE} \cite{song2024morphvae}: MorphVAE is a probabilistic generative approach specifically devised for VSR design, also using the variational autoencoder and based on EvoGym. However, different from RoboLDA, MorphVAE does not assume any hierarchical structures within its generative process, thus providing direct validation of RoboLDA's effectiveness in hierarchical modeling. 

\item \textbf{Conditional Diffusion Model (c-DM)}: c-DM is inspired by \citet{xu2024dynamics} and uses a diffusion model conditional on tasks to model high-performing robot distributions. However, we find it infeasible to implement the dynamics model in \cite{xu2024dynamics} on VSRs, due to more complex task settings and the discrete nature of voxel placement. Hence, we exclude gradient guidance from this baseline algorithm.  

\item \textbf{LASeR} \cite{song2025laser}: LASeR is one of the latest robot design algorithms that uses pretrained large language models (LLMs) to generate robot morphologies. It leverages the in-context learning ability of LLMs to analyze high-performing robot designs and propose new ones through prompting. 

\item \textbf{POET} (Paired Open-Ended Trailblazer; \citealp{wang2019paired}): POET is a co-evolutionary algorithm that simultaneously evolves agents and their environments in a multi-task setting. In our context, we adapt it as a multi-task evolutionary baseline: morphologies are co-evolved across multiple training tasks with cross-task transfer of the best individuals, and the resulting morphologies are then evaluated on unseen target tasks. This allows POET to exploit shared structure across training tasks through evolutionary transfer.
\end{itemize}

Although RoboGAN, MorphVAE and c-DM utilize generative models trained from scratch as in RoboLDA, they do not account for hierarchical structures in their architectural design. Hence, drawing comparison with these baselines can directly reveal the advantage of RoboLDA in better capturing the generative mechanisms underlying robot morphology. We fit these models using \emph{exactly the same} robot samples and aligned compute (number of parameters and training steps), and draw comparisons with their zero-shot generation capabilities. The implementation of evolutionary algorithms (EAs) (\emph{i.e., }BO, GA and CPPN-NEAT) strictly follows \citet{bhatia2021evolution} based on bi-level optimization. We compare with not only zero-shot performance of these EAs, but also their evolved results to fully showcase RoboLDA's remarkable zero-shot generalizability. 

\section{Visualization of rigid robots generated by different algorithms}

The rigid robot morphologies generated by various algorithms, as described in Appendix F of main text, are visualized in Figure \ref{fig:morphs} along with their performances in the test task. 

\begin{figure*}[h]
    \centering
    \includegraphics[width=0.65\linewidth]{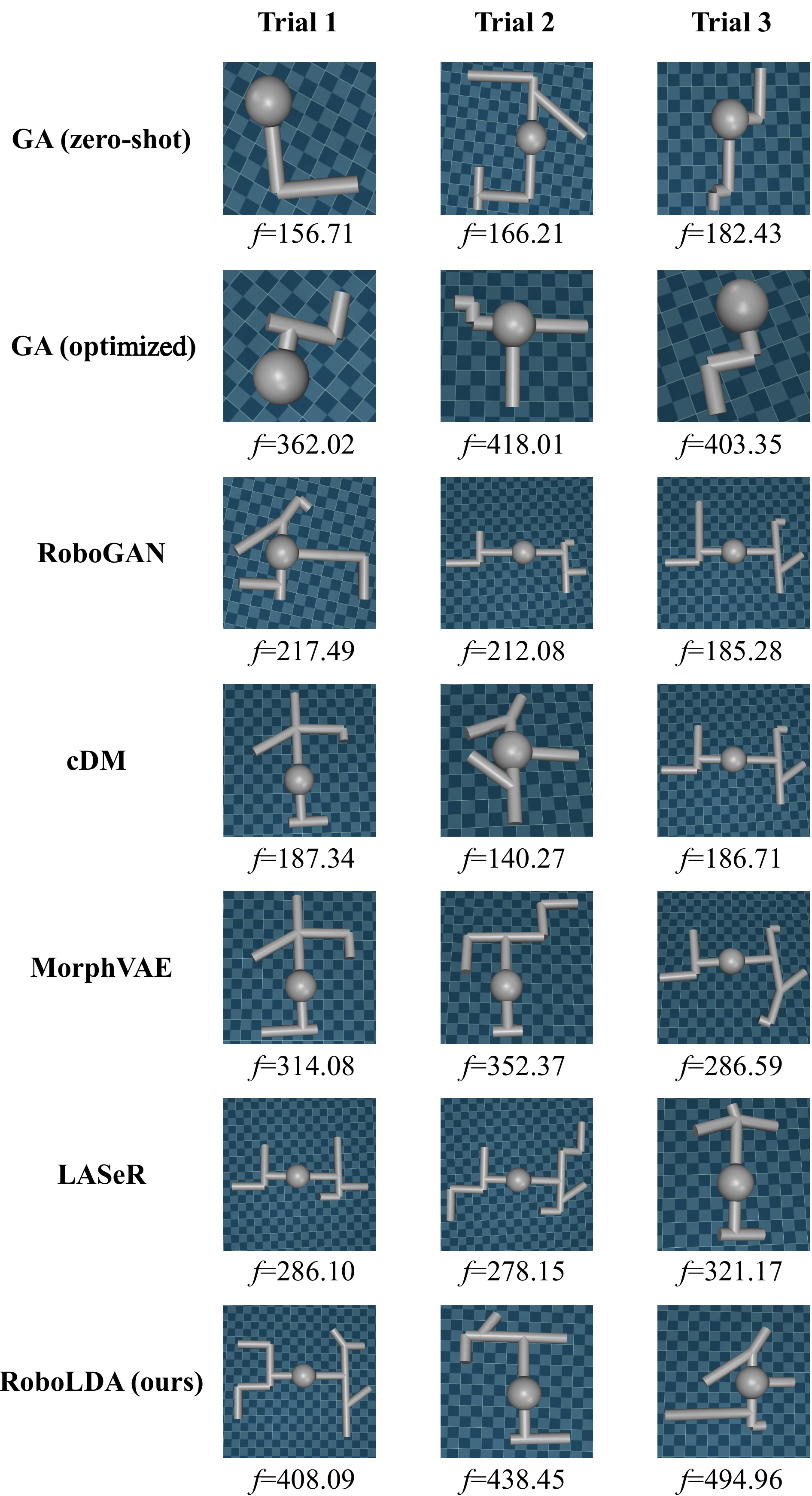}
   \caption{Visualization of the best-performing rigid robot morphologies generated by different algorithms in each trial, as well as the ones optimized by the genetic algorithm. $f$ denotes fitness of each robot, which is measured as the cumulative sum of rewards over a complete episode of 400 timesteps, when controlled by the optimized controller. Reward of each step is in turn assigned according to the velocity of the robot along the $x$-axis. }
    \label{fig:morphs}
\end{figure*}

\section{Additional visualization of hierarchical structures}

\begin{figure*}[h]
	\centering  
	\subfigbottomskip=2pt 
	\subfigcapskip=-5pt 
        \subfigure[Walker-v0 \& DownStepper-v0]{
		\includegraphics[width=0.7\linewidth]{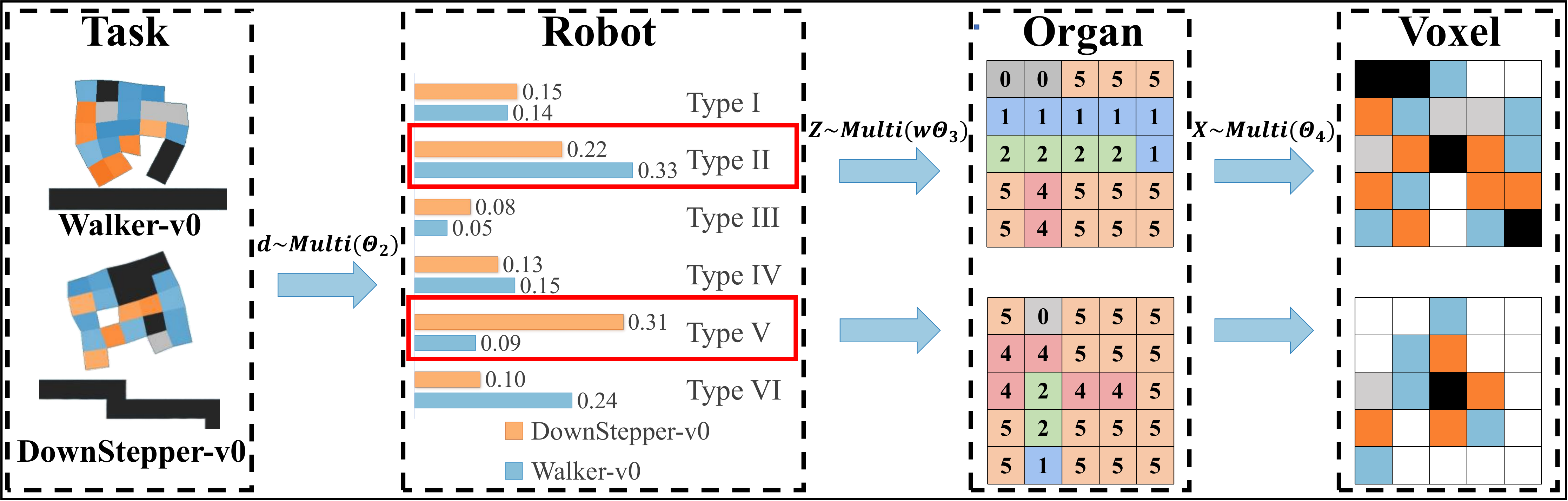}}
        \subfigure[Walker-v0 \& Pusher-v0]{
		\includegraphics[width=0.7\linewidth]{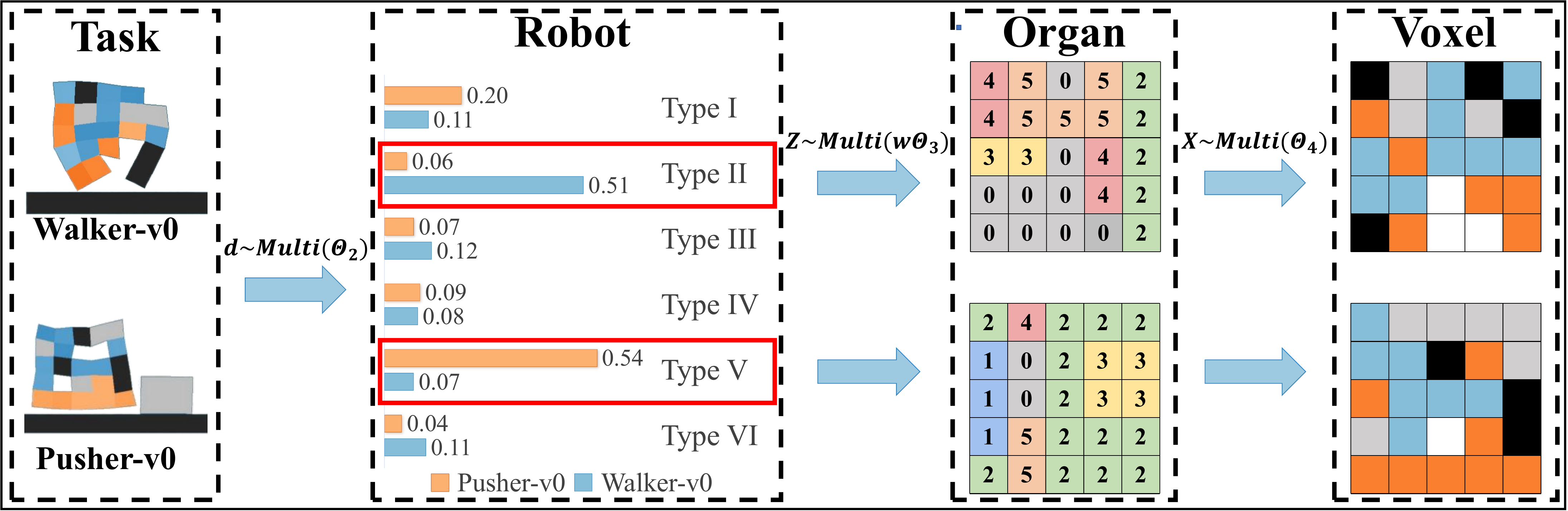}}
        \subfigure[Climber-v0]{
		\includegraphics[width=0.7\linewidth]{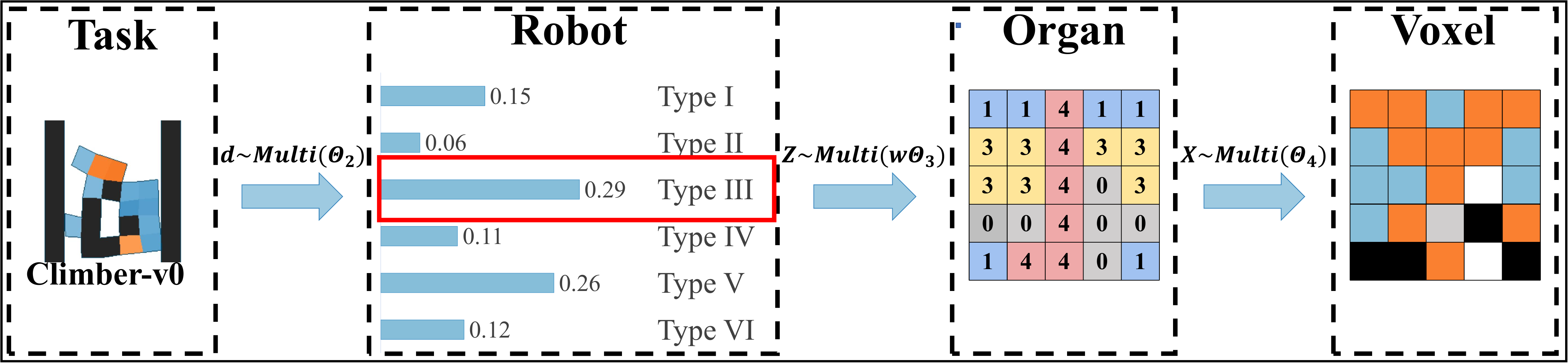}}
        \subfigure[BridgeWalker-v0]{
		\includegraphics[width=0.7\linewidth]{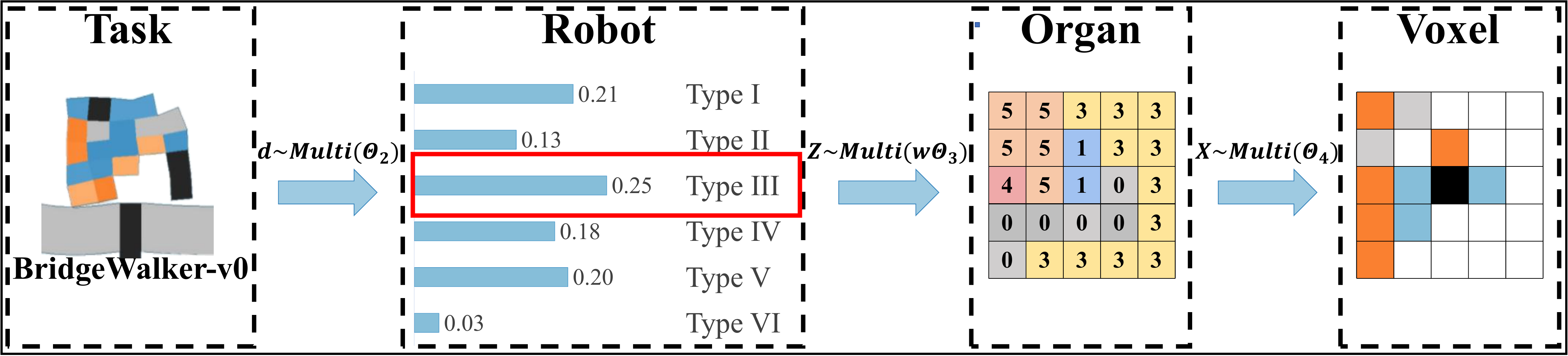}}
  \caption{Additional visualizations of hierarchical structures learned by RoboLDA. }
    \label{fig:hierarchies}
\end{figure*}

Here we showcase the hierarchical structures learned by RoboLDA in some additional task settings (Figure \ref{fig:hierarchies}). For each task combination, the most likely organ and voxel layout of the dominating robot type is illustrated. In general, the distribution of organ types exhibits significant clustering and corresponds to the functional substructures within robot morphology. Nevertheless, due to the highly abstract nature of soft robot design, the hierarchical structures are not always perfectly interpretable. We believe this is a common limitation of data-driven and deep learning approaches. It would be an exciting future direction to investigate how RoboLDA could better aid understanding and facilitate a synergistic interplay between human intuition and data-driven insights. 

\clearpage

\bibliographystyle{elsarticle-harv}
\bibliography{reference}

\end{document}